\documentclass{article}

\PassOptionsToPackage{numbers, sort, compress}{natbib}

\usepackage[preprint]{neurips_2022}

\usepackage[utf8]{inputenc} 
\usepackage[T1]{fontenc}    
\usepackage{hyperref}       
\usepackage{url}            
\usepackage{booktabs}       
\usepackage{amsfonts}       
\usepackage{nicefrac}       
\usepackage{microtype}      
\usepackage{xcolor}         

\usepackage{amsmath,amsfonts,bm}

\def\eqref#1{equation~\ref{#1}}

\def\1{\bm{1}}

\def\vc{{\bm{c}}}

\def\vv{{\bm{v}}}

\def\vx{{\bm{x}}}

\def\evc{{c}}

\def\evx{{x}}

\def\mA{{\bm{A}}}
\def\mB{{\bm{B}}}

\def\mE{{\bm{E}}}

\def\mS{{\bm{S}}}

\def\mV{{\bm{V}}}

\def\mX{{\bm{X}}}

\def\mZ{{\bm{Z}}}

\def\mPhi{{\bm{\Phi}}}

\DeclareMathAlphabet{\mathsfit}{\encodingdefault}{\sfdefault}{m}{sl}
\SetMathAlphabet{\mathsfit}{bold}{\encodingdefault}{\sfdefault}{bx}{n}

\def\sC{{\mathbb{C}}}

\newcommand{\R}{\mathbb{R}}

\newcommand{\sigmoid}{\sigma}

\usepackage{caption}
\usepackage{subcaption}
\usepackage{graphicx}
\usepackage[capitalize,noabbrev]{cleveref}
\usepackage{multirow}
\usepackage{rotating}
\usepackage{float}
\usepackage{breqn}
\usepackage{algorithm}
\usepackage{algorithmic}
\usepackage{wrapfig}
\usepackage{mathtools}

\usepackage{listings}
\def\cA{\mathcal{A}}
\def\cO{\mathcal{O}}
\def\cL{\mathcal{L}}
\def\cF{\mathcal{F}}

\def\C{\mathbb{C}}

\def\valpha{\bm{\alpha}}
\def\mPhi{\bm{\Phi}}

\mathchardef\mhyphen="2D
\def\ESA{\cA_{\mathrm{ES}}}
\def\topk{\mathrm{Top \mhyphen K}}
\def\argtopk{\mathop{\arg\topk}}

\newcommand{\level}[2]{\mE_{#1}^{(#2)}}
\newcommand{\trend}[2]{\mB_{#1}^{(#2)}}
\newcommand{\season}[2]{\mS_{#1}^{(#2)}}
\newcommand{\res}[2]{\mZ_{#1}^{(#2)}}

\def\hrz{t:t+H}
\def\lb{t-L:t}

\newcommand\doublebar{\kern1pt\rule[-\dp\strutbox]{1pt}{\baselineskip}\kern1pt \kern1pt\rule[-\dp\strutbox]{1pt}{\baselineskip}\kern1pt}

\def\shortname{ETSformer}
\def\longname{Exponential Smoothing Transformers for Time-series Forecasting}

\title{{\shortname}: {\longname}}

\newcommand*{\affaddr}[1]{#1} 
\newcommand*{\affmark}[1][*]{\textsuperscript{#1}}
\newcommand*{\email}[1]{\texttt{#1}}
\usepackage{caption}
\makeatletter
\newcommand{\printfnsymbol}[1]{%
  \textsuperscript{\@fnsymbol{#1}}%
}
\makeatother

\author{%
  Gerald Woo\affmark[1] \affmark[2], 
  Chenghao Liu\affmark[1], 
  Doyen Sahoo\affmark[1], 
  Akshat Kumar\affmark[2] \&
  Steven Hoi\affmark[1] \\
  \affaddr{\affmark[1]}Salesforce Research Asia, \affaddr{\affmark[2]}Singapore Management University\\
  \email{\tt\small\{gwoo,chenghao.liu,dsahoo,shoi\}@salesforce.com}\\
  \email{\tt\small\{akshatkumar\}@smu.edu.sg}
}

\begin{document}

\maketitle

\begin{abstract}
    Transformers have been actively studied for time-series forecasting in recent years. While often showing promising results in various scenarios, traditional Transformers are not designed to fully exploit the characteristics of time-series data and thus suffer some fundamental limitations, e.g., they are generally not decomposable or interpretable, and are neither effective nor efficient for long-term forecasting. In this paper, we propose {\shortname}, a novel time-series Transformer architecture, which exploits the principle of exponential smoothing in improving Transformers for time-series forecasting.  In particular, inspired by the classical exponential smoothing methods in time-series forecasting, we propose the novel exponential smoothing attention (ESA) and frequency attention (FA) to replace the self-attention mechanism in vanilla Transformers, thus improving both accuracy and efficiency. Based on these, we redesign the Transformer architecture with modular decomposition blocks such that it can learn to decompose the time-series data into interpretable time-series components such as level, growth and seasonality. Extensive experiments on various time-series benchmarks validate the efficacy and advantages of the proposed method. Code is available at \url{https://github.com/salesforce/ETSformer}. 
\end{abstract}
\section{Introduction}
Transformer models have achieved great success in the fields of NLP \cite{vaswani2017attention, Devlin2019BERT} and CV \cite{carion2020end, dosovitskiy2021an} in recent times. The success is widely attributed to its self-attention mechanism which is able to explicitly model both short and long range dependencies adaptively via the pairwise query-key interaction. Owing to their powerful capability to model sequential data, Transformer-based architectures \cite{LI2019EnhancingTL,Wu2020AdversarialST,zhou2021informer, wu2021autoformer,zerveas2021transformer} have been actively explored for the time-series forecasting, especially for the more challenging Long Sequence Time-series Forecasting (LSTF) task. While showing promising results, it is still quite challenging to extract salient temporal patterns and thus make accurate long-term forecasts for large-scale data. This is because time-series data is usually noisy and non-stationary. Without incorporating appropriate knowledge about time-series structures \cite{assimakopoulos2000theta, theodosiou2011forecasting, hyndman2008forecasting}, it is prone to learning the spurious dependencies and lacks interpretability.

Moreover, the use of content-based, dot-product attention in Transformers is not effective in detecting essential temporal dependencies for two reasons. 
(1) Firstly, time-series data is usually assumed to be generated by a conditional distribution over past observations, with the dependence between observations weakening over time \cite{mohri2010stability, kuznetsov2015learning}. Therefore, neighboring data points have similar values, and recent tokens should be given a higher weight\footnote{An assumption further supported by the success of classical exponential smoothing methods and ARIMA model selection methods tending to select small lags.} when measuring their similarity \cite{hyndman2008forecasting, hyndman2008automatic}. This indicates that attention measured by a relative time lag is more effective than that measured by the similarity of the content when modeling time-series. 
(2) Secondly, many real world time-series display strong seasonality -- patterns in time-series which repeat with a fixed period. Automatically extracting seasonal patterns has been proved to be critical for the success of forecasting \cite{cleveland1976decomposition, cleveland1990stl, woo2022cost}.   However, the vanilla attention mechanism is unlikely able to learn these required periodic dependencies without any in-built prior structure.

\begin{wrapfigure}{r}{0.5\textwidth}
\vspace{-0.1in}
\centering
\resizebox{0.4\textwidth}{!}{
\includegraphics[width=\columnwidth]{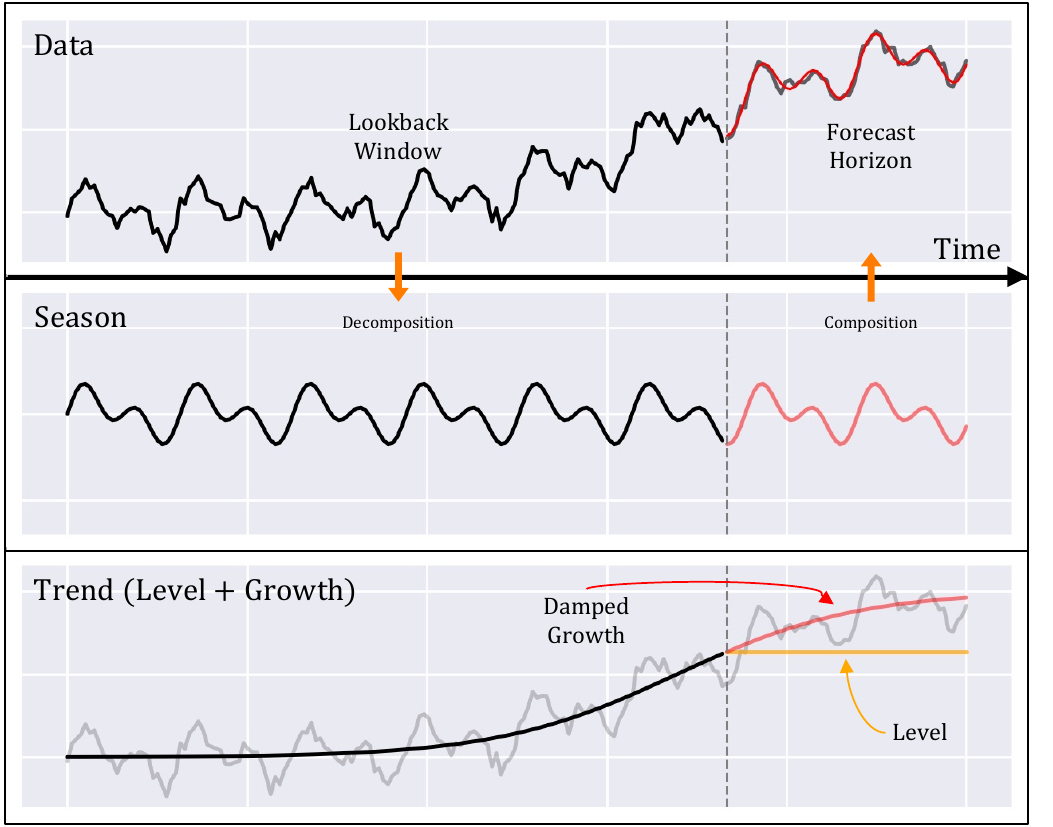}
}
\caption{Illustration demonstrating how ETSformer generates forecasts via a decomposition (of intermediate representations) into seasonal and trend components. The seasonal component extracts salient periodic patterns and extrapolates them. The trend component which is a combination of the level and growth terms, first estimates the current level of the time-series, and subsequently adds a damped growth term to generate trend forecasts.}
\label{fig:decomposition}
\end{wrapfigure}

To address these limitations, we propose {\shortname}, an effective and efficient Transformer architecture for time-series forecasting,  inspired by exponential smoothing methods \cite{hyndman2008forecasting} and illustrated in \cref{fig:decomposition}. First of all, {\shortname} incorporates inductive biases of time-series structures by performing a layer-wise level, growth, and seasonal decomposition. By leveraging the high capacities of deep architectures and an effective residual learning scheme, {\shortname} is able to extract a series of latent growth and seasonal patterns and model their complex dependencies.
Secondly, {\shortname} introduces a novel Exponential Smoothing Attention (ESA) and Frequency Attention (FA) to replace vanilla attention. In particular, ESA constructs attention scores based on the relative time lag to the query, and achieves \(\cO(L \log L)\) complexity for the length-$L$ lookback window and demonstrates powerful capability in modeling the growth component. FA leverages the Fourier transformation to extract the dominating seasonal patterns by selecting the Fourier bases with the \(K\) largest amplitudes in frequency domain, and also achieves \(\cO(L \log L)\) complexity. Finally, the predicted forecast is a composition of level, trend, and seasonal components, which makes it human interpretable. We conduct extensive empirical analysis and show that {\shortname} achieves state-of-the-art performance by outperforming competing approaches over 6 real world datasets on both the multivariate and univariate settings, and also visualize the time-series components to verify its interpretability.
\section{Related Work}
\textbf{Transformer based deep forecasting.} Inspired by the success of Transformers in CV and NLP, Transformer-based time-series forecasting models have been actively studied recently. LogTrans \cite{LI2019EnhancingTL} introduces local context to Transformer models via causal convolutions in the query-key projection layer, and propose the LogSparse attention to reduce complexity to \(\cO(L \log L)\). Informer \cite{zhou2021informer} extends the Transformer by proposing the ProbSparse attention and distillation operation to achieve \(\cO(L \log L)\) complexity. AST \cite{Wu2020AdversarialST} leverages a sparse normalization transform, \(\alpha\mhyphen\mathrm{entmax}\), to implement a sparse attention layer. It further incorporates an adversarial loss to mitigate the adverse effect of error accumulation in inference. Similar to our work that incorporates prior knowledge of time-series structure, Autoformer \cite{wu2021autoformer} introduces the Auto-Correlation attention mechanism which focuses on sub-series based similarity and is able to extract periodic patterns. 
Yet, their implementation of series decomposition which performs de-trending via a simple moving average over the input signal without any learnable parameters is arguably a simplified assumption, insufficient to appropriately model complex trend patterns. {\shortname} on the other hand, decomposes the series by de-seasonalization as seasonal patterns are more identifiable and easier to detect \cite{de2011forecasting}. Furthermore, the Auto-Correlation mechanism fails to attend to information from the local context (i.e. forecast at \(t+1\) is not dependent on \(t, t-1\), etc.) and does not separate the trend component into level and growth components, which are both crucial for modeling trend patterns. Lastly, similar to previous work, their approach is highly reliant on manually designed dynamic time-dependent covariates (e.g. month-of-year, day-of-week), while {\shortname} is able to automatically learn and extract seasonal patterns from the time-series signal directly.

\textbf{Attention Mechanism.} 
The self-attention mechanism in Transformer models has recently received much attention, its necessity has been greatly investigated in attempts to introduce more flexibility and reduce computational cost. Synthesizer \cite{tay2021synthesizer} empirically studies the importance of dot-product interactions, and show that a randomly initialized, learnable attention mechanisms with or without token-token dependencies can achieve competitive performance with vanilla self-attention on various NLP tasks. \cite{you2020hard} utilizes an unparameterized Gaussian distribution to replace the original attention scores, concluding that the attention distribution should focus on a certain local window and can achieve comparable performance. \cite{raganato2020fixed} replaces attention with fixed, non-learnable positional patterns, obtaining competitive performance on NMT tasks. \cite{lee2021fnet} replaces self-attention with a non-learnable Fourier Transform and verifies it to be an effective mixing mechanism. While our proposed ESA shares the spirit of designing attention mechanisms that are not dependent on pair-wise query-key interactions, our work is inspired by exploiting the characteristics of time-series and is an early attempt to utilize prior knowledge of time-series for tackling the time-series forecasting tasks.
\section{Preliminaries and Background}
\textbf{Problem Formulation} Let \(\vx_t \in \R^m\) denote an observation of a multivariate time-series at time step \(t\). Given a lookback window \(\mX_{\lb} = [\vx_{t-L}, \ldots, \vx_{t-1}]\), we consider the task of predicting future values over a horizon, \(\mX_{\hrz} = [\vx_t, \ldots, \vx_{t+H-1}]\). We denote \(\hat{\mX}_{\hrz}\) as the point forecast of \(\mX_{\hrz}\). Thus, the goal is to learn a forecasting function \(\hat{\mX}_{\hrz} = f (\mX_{\lb})\) by minimizing some loss function \(\cL: \R^{H \times m} \times \R^{H \times m} \to \R\).

\textbf{Exponential Smoothing} We instantiate exponential smoothing methods \cite{hyndman2008forecasting} in the univariate forecasting setting. They assume that time-series can be decomposed into seasonal and trend components, and trend can be further decomposed into level and growth components. Specifically, a commonly used model is the additive Holt-Winters' method \cite{holt2004forecasting, winters1960forecasting}, which can be formulated as:
\begin{align}
\label{eq:es}
\text{Level}&:  e_{t} =  \alpha(x_t - s_{t-p}) + (1-\alpha)(e_{t-1}+b_{t-1})\nonumber\\
\text{Growth}&:  b_{t} = \beta(e_{t} - e_{t-1}) + (1-\beta)b_{t-1}\nonumber\\
\text{Seasonal}&:  s_t = \gamma(x_t - e_t) + (1-\gamma)s_{t-p}\nonumber\\
\text{Forecasting}&:\hat{x}_{t+h|t} = e_t + hb_t +s_{t+h - p}
\end{align}

where $p$ is the period of seasonality, and $\hat{x}_{t+h|t}$ is the $h$-steps ahead forecast. The above equations state that the $h$-steps ahead forecast is composed of the last estimated level $e_t$, incrementing it by $h$ times the last growth factor, $b_t$, and adding the last available seasonal factor $s_{t+h-p}$. Specifically, the level smoothing equation is formulated as a weighted average of the seasonally adjusted observation $(x_t- s_{t-p})$ and the non-seasonal forecast, obtained by summing the previous level and growth $(e_{t-1} + b_{t-1})$. The growth smoothing equation is implemented by a weighted average between the successive difference of the (de-seasonalized) level, $(e_t - e_{t-1})$, and the previous growth, $b_{t-1}$. Finally, the seasonal smoothing equation is a weighted average between the difference of observation and (de-seasonalized) level, $(x_t - e_t)$, and the previous seasonal index $s_{t-p}$. The weighted average of these three equations are controlled by the smoothing parameters $\alpha$, $\beta$ and $\gamma$, respectively. 

A widely used modification of the additive Holt-Winters' method is to allow the damping of trends, which has been proved to produce robust multi-step forecasts \cite{svetunkov2016complex, mckenzie2010damped}. The forecast with damping trend can be rewritten as:
\begin{align}
\label{eq:damping}
\hat{x}_{t+h|t} = e_t + (\phi+\phi^2+\dots+\phi^h)b_t +s_{t+h - p},    
\end{align}
where the growth is damped by a factor of $\phi$. If $\phi=1$, it degenerates to the vanilla forecast. For $0<\phi<1$, as $h \rightarrow \infty$ this growth component approaches an asymptote given by $\phi b_t/(1- \phi)$.  
\section{{\shortname}}
\begin{figure*}[t]
\begin{center}
\centerline{\includegraphics[width=\textwidth]{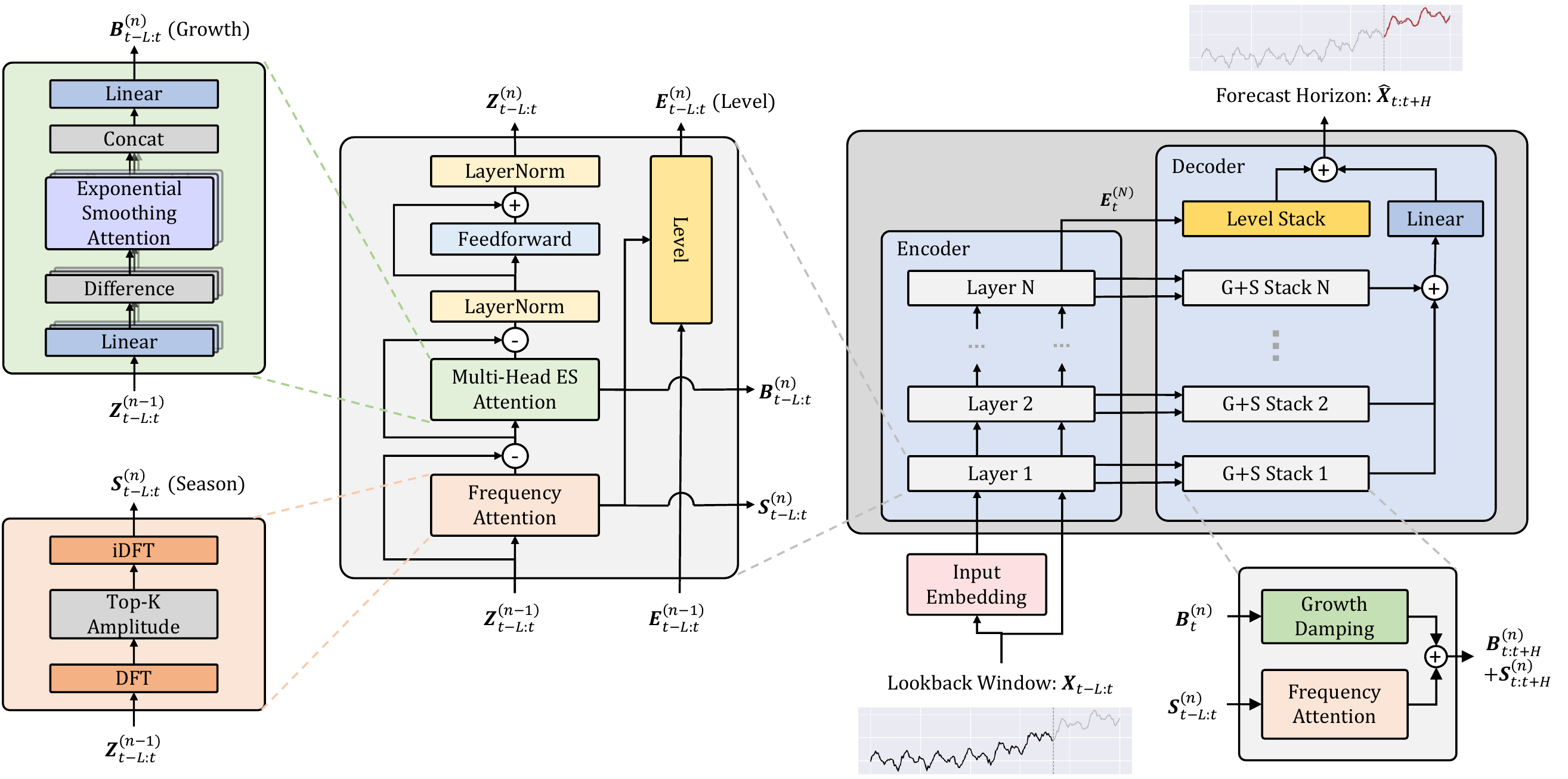}}
\caption{ETSformer model architecture.}
\label{fig:overall-architecture}
\end{center}
\vskip -0.3in
\end{figure*}
In this section, we redesign the classical Transformer architecture into an exponential smoothing inspired encoder-decoder architecture specialized for tackling the time-series forecasting problem.
Our architecture design methodology relies on three key principles: 
(1) the architecture leverages the stacking of multiple layers to progressively extract a series of level, growth, and seasonal representations from the intermediate latent residual;
(2) following the spirit of exponential smoothing, we extract the salient seasonal patterns while modeling level and growth components by assigning higher weight to recent observations; 
(3) the final forecast is a composition of level, growth, and seasonal components making it human interpretable. 
We now expound how our {\shortname} architecture encompasses these principles.

\subsection{Overall Architecture}
\cref{fig:overall-architecture} illustrates the overall encoder-decoder architecture of {\shortname}. At each layer, the encoder is designed to iteratively extract growth and seasonal latent components from the lookback window. 
The level is then extracted in a similar fashion to classical level smoothing in \cref{eq:es}.
These extracted components are then fed to the decoder to further generate the final $H$-step ahead forecast via a composition of level, growth, and seasonal forecasts, which is defined:

\vskip -0.2in
\begin{small}
\begin{align}
\label{eq:forecast}
\hat{\mX}_{\hrz}  = \mE_{\hrz}  + \mathrm{Linear} \Big(  \sum^N_{n=1}(\trend{\hrz}{n}
+ \season{\hrz}{n})\Big), 
\end{align}
\end{small}
\vskip -0.1in

where \(\mE_{\hrz} \in \R^{H \times m}\), and \(\trend{\hrz}{n}, \season{\hrz}{n} \in \R^{H \times d}\) represent the level forecasts, and the growth and seasonal latent representations of each time step in the forecast horizon, respectively. The superscript represents the stack index, for a total of \(N\) encoder stacks. Note that \(\mathrm{Linear}(\cdot): \R^d \to \R^m\) operates element-wise along each time step, projecting the extracted growth and seasonal representations from latent to observation space.

\subsubsection{Input Embedding}
Raw signals from the lookback window are mapped to latent space via the input embedding module, defined by \(\res{\lb}{0} = \level{\lb}{0} = \mathrm{Conv}(\mX_{\lb}),\)
where \(\mathrm{Conv}\) is a temporal convolutional filter with kernel size 3, input channel \(m\) and output channel \(d\). In contrast to prior work \cite{LI2019EnhancingTL, Wu2020AdversarialST, wu2021autoformer,zhou2021informer}, the inputs of {\shortname} do not rely on any other manually designed dynamic time-dependent covariates (e.g. month-of-year, day-of-week) for both the lookback window and forecast horizon. This is because the proposed Frequency Attention module (details in \cref{subsubsec:fa}) is able to automatically uncover these seasonal patterns, which renders it more applicable for challenging scenarios without these discriminative covariates and reduces the need for feature engineering. 

\subsubsection{Encoder}
The encoder focuses on extracting a series of latent growth and seasonality representations in a cascaded manner from the lookback window.
To achieve this goal, traditional methods rely on the assumption of additive or multiplicative seasonality which has limited capability to express complex patterns beyond these assumptions. Inspired by \cite{oreshkin2019n, he2016deep}, we leverage residual learning to build an expressive, deep architecture to characterize the complex intrinsic patterns. Each layer can be interpreted as sequentially analyzing the input signals. The extracted growth and seasonal signals are then removed from the residual and undergo a nonlinear transformation before moving to the next layer. Each encoder layer takes as input the residual from the previous encoder layer \(\res{\lb}{n-1}\) and emits \(\res{\lb}{n}, \trend{\lb}{n}, \season{\lb}{n}\), the residual, latent growth, and seasonal representations for the lookback window via the Multi-Head Exponential Smoothing Attention (MH-ESA) and Frequency Attention (FA) modules (detailed description in \cref{subsec:esa-and-fa}). The following equations formalizes the overall pipeline in each encoder layer, and for ease of exposition, we use the notation \(\vcentcolon =\) for a variable update.
\noindent\begin{minipage}[t]{.5\linewidth}
\begin{align*}
    \textrm{Seasonal:} \quad
    \season{\lb}{n} & = \mathrm{FA}_{\lb}(\res{\lb}{n-1}) \\
    \res{\lb}{n-1} & \vcentcolon = \res{\lb}{n-1} - \season{\lb}{n}
\end{align*}
\end{minipage}%
\begin{minipage}[t]{.5\linewidth}
\begin{align*}
    \textrm{Growth:} \quad
    \trend{\lb}{n} & = \mathrm{MH}\mhyphen\mathrm{ESA}(\res{\lb}{n-1}) \\
    \res{\lb}{n-1} & \vcentcolon = \mathrm{LN}(\res{\lb}{n-1} - \trend{\lb}{n}) \\
    \res{\lb}{n} & = \mathrm{LN}(\res{\lb}{n-1} + \mathrm{FF}(\res{\lb}{n-1}))
\end{align*}
\end{minipage}

\(\mathrm{LN}\) is layer normalization \cite{ba2016layer},  \(\mathrm{FF}(x) = \mathrm{Linear}(\sigmoid(\mathrm{Linear}(x)))\) is a position-wise feedforward network \cite{vaswani2017attention} and $\sigmoid(\cdot)$ is the sigmoid function.

\textbf{Level Module} Given the latent growth and seasonal representations from each layer, we extract the level at each time step $t$ in the lookback window in a similar way as the level smoothing equation in \cref{eq:es}. Formally, the adjusted level is a weighted average of the current (de-seasonalized) level and the level-growth forecast from the previous time step $t-1$. It can be formulated as:
\begin{align*}
    \level{t}{n} = \valpha * \Big( \level{t}{n-1} - \mathrm{Linear}(\season{t}{n}) \Big)
    + (1 - \valpha) * \Big( \level{t-1}{n} + \mathrm{Linear}(\trend{t-1}{n}) \Big),
\end{align*}
where \(\valpha \in \R^m\) is a learnable smoothing parameter, \(*\) is an element-wise multiplication term, and \(\mathrm{Linear}(\cdot): \R^d \to \R^m\) maps representations to observation space. Finally, the extracted level in the last layer $\level{\lb}{N}$ can be regarded as the corresponding level for the lookback window.
We show in \cref{subapp:level-esa} that this recurrent exponential smoothing equation can also be efficiently evaluated using the efficient \(\ESA\) algorithm (\cref{alg:efficient-esa}) with an auxiliary term.

\subsubsection{Decoder}
The decoder is tasked with generating the $H$-step ahead forecasts. As shown in \cref{eq:forecast}, the final forecast is a composition of level forecasts $\mE_{\hrz}$, growth representations $\trend{\hrz}{n}$ and seasonal representations $\season{\hrz}{n}$ in the forecast horizon. It comprises \(N\) Growth + Seasonal (G+S) Stacks, and a Level Stack. The G+S Stack consists of the Growth Damping (GD) and FA blocks, which leverage \(\trend{t}{n}\), \(\season{\lb}{n}\) to predict \(\trend{\hrz}{n}\), \(\season{\hrz}{n}\), respectively.
\noindent\begin{minipage}[t]{.5\textwidth}
\begin{align*}
    \text{Growth:}\quad\trend{\hrz}{n} = \mathrm{TD}(\trend{t}{n})
\end{align*}
\end{minipage}%
\begin{minipage}[t]{.5\textwidth}
\begin{align*}
    \text{Seasonal:}\quad\season{\hrz}{n} = \mathrm{FA}_{\hrz}(\season{\lb}{n})
\end{align*}
\end{minipage}

To obtain the level in the forecast horizon, the Level Stack repeats the level in the last time step $t$ along the forecast horizon. It can be defined as $\mE_{\hrz} = \mathrm{Repeat}_H(\level{t}{N}) = [\level{t}{N}, \ldots, \level{t}{N}]$, with $\mathrm{Repeat}_H(\cdot): \R^{1 \times m} \to \R^{H\times m}$.

\textbf{Growth Damping} To obtain the growth representation in the forecast horizon, we follow the idea of trend damping in \cref{eq:damping} to make robust multi-step forecast.
Thus, the trend representations can be formulated as:
\begin{align*}
    \mathrm{TD}(\trend{t}{n})_j & = \sum_{i=1}^{j} \gamma^i \trend{t}{n}, \\
    \mathrm{TD}(\trend{\lb}{n}) & = [\mathrm{TD}(\trend{t}{n})_t, \ldots, \mathrm{TD}(\trend{t}{n})_{t+H-1}],
\end{align*}
where \(0 < \gamma < 1\) is the damping parameter which is learnable, and in practice, we apply a multi-head version of trend damping by making use of \(n_h\) damping parameters. Similar to the implementation for level forecast in the Level Stack, we only use the last trend representation in the lookback window $\trend{t}{n}$ to forecast the trend representation in the forecast horizon. 

\subsection{Exponential Smoothing Attention and Frequency Attention Mechanism}
\label{subsec:esa-and-fa}
\begin{figure*}[t]
\centering
\begin{subfigure}[t]{.32\textwidth}
    \centering
    \includegraphics[width=\textwidth]{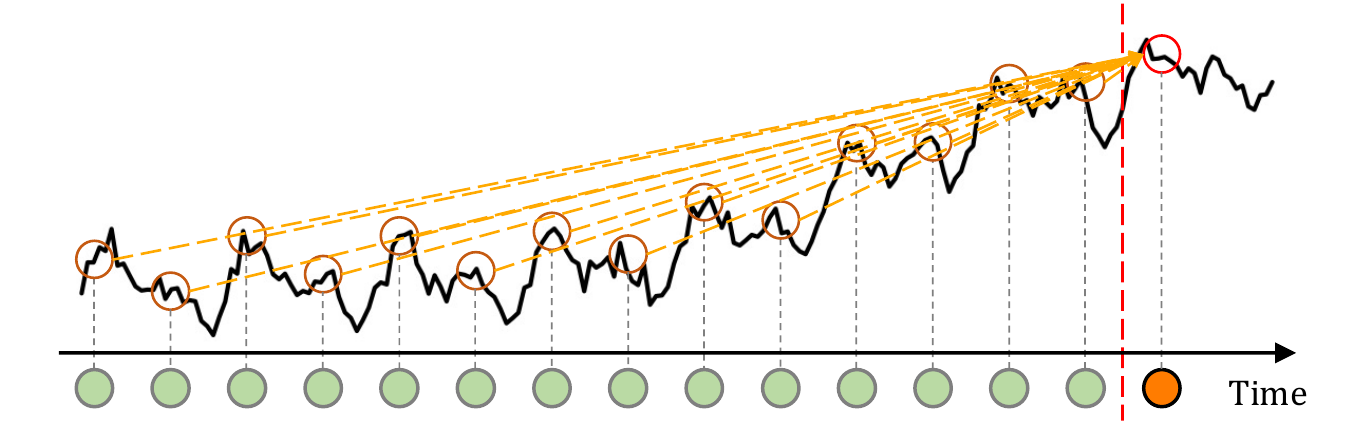}
    \caption{Full Attention (\citeyear{vaswani2017attention})}
    \label{subfig:full}
\end{subfigure}
\begin{subfigure}[t]{.32\textwidth}
    \centering
    \includegraphics[width=\textwidth]{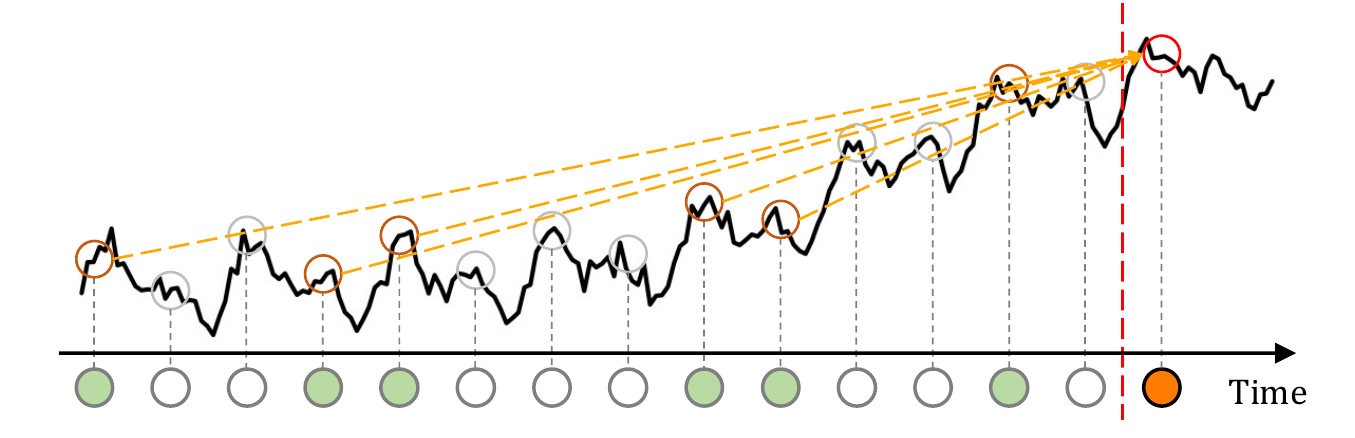}
    \caption{Sparse Attention (\citeyear{Kitaev2020Reformer, zhou2021informer})}
    \label{subfig:sparse}
\end{subfigure}
\begin{subfigure}[t]{.32\textwidth}
    \centering
    \includegraphics[width=\textwidth]{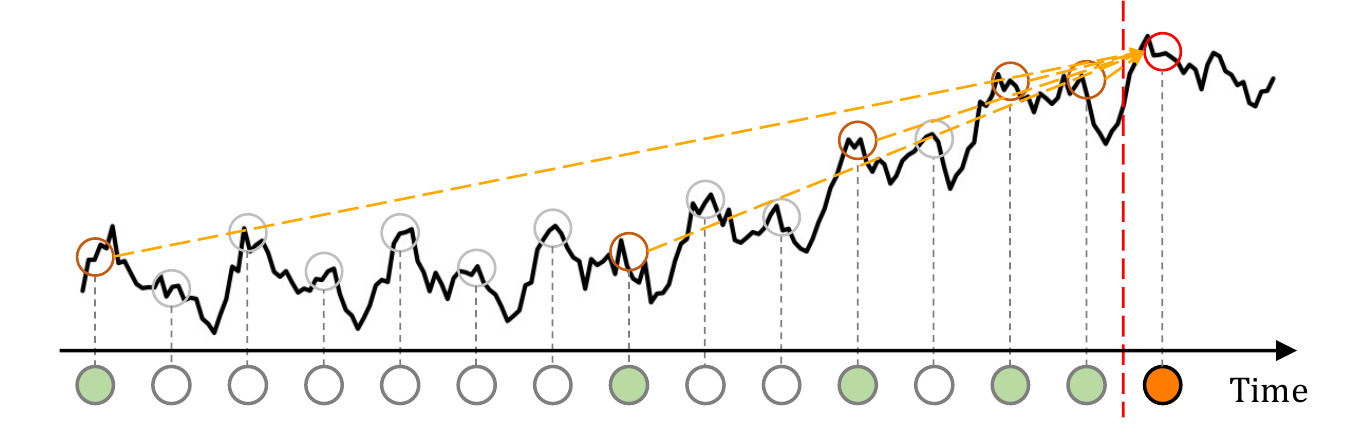}
    \caption{Log-sparse Attention (\citeyear{LI2019EnhancingTL})}
    \label{subfig:logsparse}
\end{subfigure}
\begin{subfigure}[t]{.32\textwidth}
    \centering
    \includegraphics[width=\textwidth]{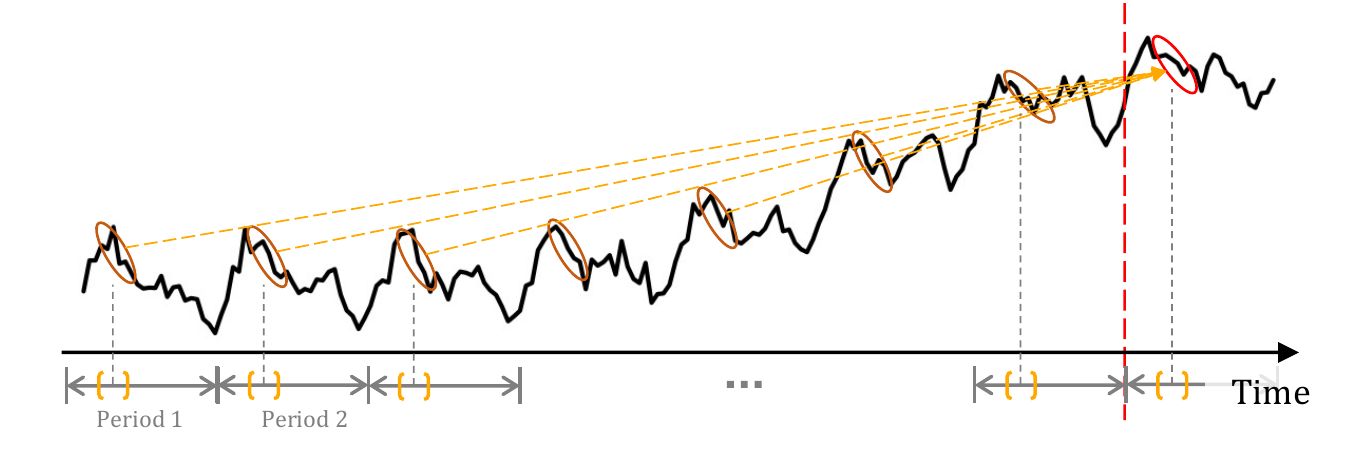}
    \caption{Autocorrelation Attention (\citeyear{wu2021autoformer})}
    \label{subfig:auto}
\end{subfigure}
\begin{subfigure}[t]{.32\textwidth}
    \centering
    \includegraphics[width=\textwidth]{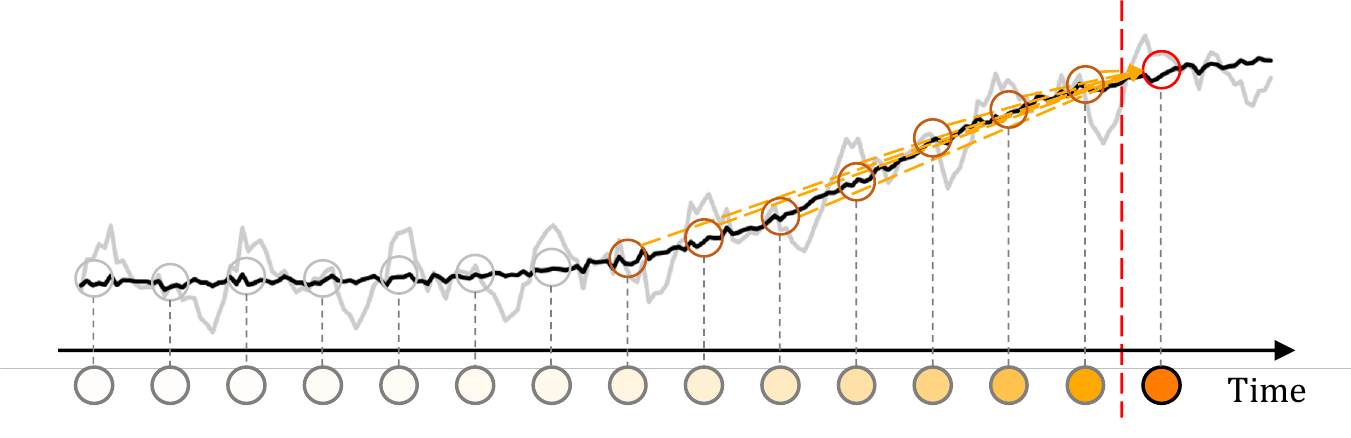}
    \caption{Exponential Smoothing Attention (Ours)}
    \label{subfig:es}
\end{subfigure}
\begin{subfigure}[t]{.32\textwidth}
    \centering
    \includegraphics[width=\textwidth]{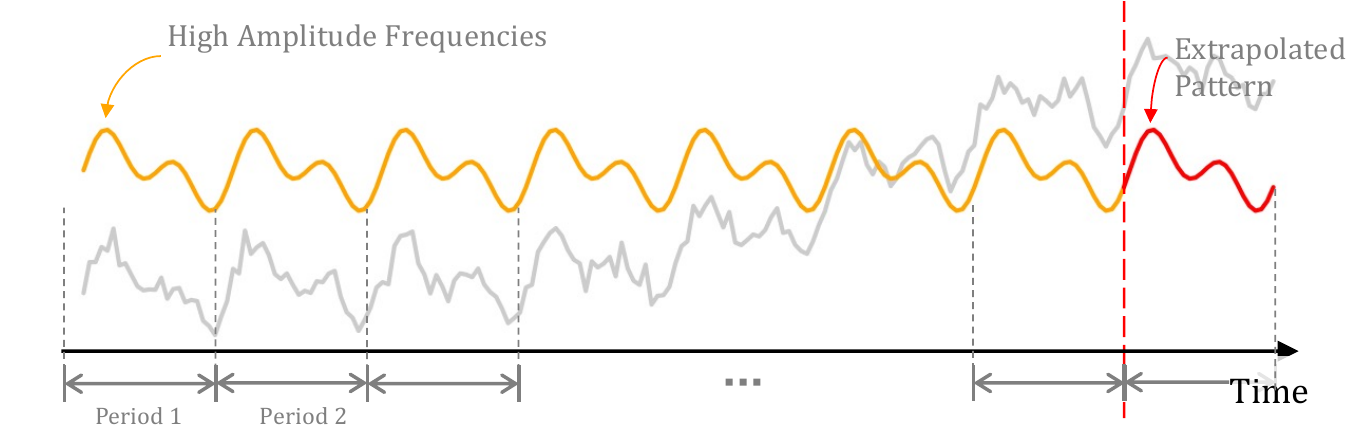}
    \caption{Frequency Attention (Ours)}
    \label{subfig:freq}
\end{subfigure}
\caption{Comparison between different attention mechanisms. (a) Full, (b) Sparse, and (c) Log-sparse Attentions are adaptive mechanisms, where the green circles represent the attention weights adaptively calculated by a point-wise dot-product query, and depends on various factors including the time-series value, additional covariates (e.g. positional encodings, time features, etc.). (d) Autocorrelation attention considers sliding dot-product queries to construct attention weights for each rolled input series.
We introduce (e) Exponential Smoothing Attention (ESA) and (f) Frequency Attention (FA). ESA directly computes attention weights based on the relative time lag, without considering the input content, while FA attends to patterns which dominate with large magnitudes in the frequency domain.}
\label{fig:attention-mechanisms}
\vskip -0.05in
\end{figure*}
Considering the ineffectiveness of existing attention mechanisms in handling time-series data, we develop the Exponential Smoothing Attention (ESA) and Frequency Attention (FA) mechanisms to extract latent growth and seasonal representations. ESA is a non-adaptive, learnable attention scheme with an inductive bias to attend more strongly to recent observations by following an exponential decay, while FA is a non-learnable attention scheme, that leverages Fourier transformation to select dominating seasonal patterns. A comparison between existing work and our proposed ESA and FA is illustrated in \cref{fig:attention-mechanisms}.

\subsubsection{Exponential Smoothing Attention}
Vanilla self-attention can be regarded as a weighted combination of an input sequence, where the weights are normalized alignment scores measuring the similarity between input contents \cite{tsai2019transformer}.
Inspired by the exponential smoothing in \cref{eq:es}, we aim to assign a higher weight to recent observations. It can be regarded as a novel form of attention whose weights are computed by the relative time lag, rather than input content.  
Thus, the ESA mechanism can be defined as \(\ESA: \R^{L \times d} \to \R^{L \times d}\), where \(\ESA(\mV)_t \in \R^d\) denotes the \(t\)-th row of the output matrix, representing the token corresponding to the \(t\)-th time step. Its exponential smoothing formula can be further written as:
\begin{align*}
\ESA(\mV)_t &= \alpha\mV_{t} +  (1-\alpha)\ESA(\mV)_{t-1} = \sum_{j=0}^{t-1} \alpha (1 - \alpha)^j \mV_{t-j} + (1 - \alpha)^t \vv_0,
\end{align*}
where \(0 < \alpha < 1\) and \(\vv_0\) are learnable parameters known as the smoothing parameter and initial state respectively. 

\textbf{Efficient $\ESA$ algorithm}
The straightforward implementation of the ESA mechanism by constructing the attention matrix, \(\mA_{\mathrm{ES}}\) and performing a matrix multiplication with the input sequence (detailed algorithm in \cref{subapp:esa-implementation}) results in an \(\cO(L^2)\) computational complexity.

\begin{align*}
    \ESA(\mV) = 
    \begin{bmatrix}
        \ESA(\mV)_1 \\
        \vdots \\
        \ESA(\mV)_L
    \end{bmatrix}
    = \mA_{\mathrm{ES}} \cdot
    \begin{bmatrix}
        \vv_0^T \\ 
        \mV
    \end{bmatrix},
\end{align*}

Yet, we are able to achieve an efficient algorithm by exploiting the unique structure of the exponential smoothing attention matrix, \(\mA_{\mathrm{ES}}\), which is illustrated in \cref{subapp:esa-matrix}. Each row of the attention matrix can be regarded as iteratively right shifting with padding (ignoring the first column). Thus, a matrix-vector multiplication can be computed with a cross-correlation operation, which in turn has an efficient fast Fourier transform implementation \cite{Mathieu2014FastTO}. The full algorithm is described in \cref{alg:efficient-esa}, \cref{subapp:esa-algo}, achieving an \(\cO(L \log L)\) complexity.

\textbf{Multi-Head Exponential Smoothing Attention (MH-ESA)}  We use \(\ESA\) as a basic building block, and develop the Multi-Head Exponential Smoothing Attention to extract latent growth representations. Formally, we obtain the growth representations by taking the successive difference of the residuals.
\begin{align*}
    \tilde{\mZ}_{\lb}^{(n)} & = \mathrm{Linear}(\res{\lb}{n-1}), \\
    \trend{\lb}{n} & = \mathrm{MH}\mhyphen\ESA(\tilde{\mZ}_{\lb}^{(n)} - [\tilde{\mZ}_{\lb-1}^{(n)}, \vv_0^{(n)}]), \\
    \trend{\lb}{n} & \coloneqq \mathrm{Linear}(\trend{\lb}{n}),
\end{align*}
where \(\mathrm{MH}\mhyphen\ESA\) is a multi-head version of \(\ESA\) and \(\vv_0^{(n)}\) is the initial state from the ESA mechanism.

\subsubsection{Frequency Attention}
\label{subsubsec:fa}
The goal of identifying and extracting seasonal patterns from the lookback window is twofold. Firstly, it can be used to perform de-seasonalization on the input signals such that downstream components are able to focus on modeling the level and growth information. Secondly, we are able to extrapolate the seasonal patterns to build representations for the forecast horizon. 
The main challenge is to automatically identify seasonal patterns. Fortunately, the use of power spectral density estimation for periodicity detection has been well studied \cite{vlachos2005periodicity}. Inspired by these methods, we leverage the discrete Fourier transform (DFT, details in \cref{app:dft}) to develop the FA mechanism  to extract dominant seasonal patterns.

Specifically, FA first decomposes input signals into their Fourier bases via a DFT along the temporal dimension, \(\cF(\res{\lb}{n-1}) \in \C^{F \times d}\) where \(F = \lfloor L /2 \rfloor +1\), and selects bases with the \(K\) largest amplitudes. An inverse DFT is then applied to obtain the seasonality pattern in time domain.
Formally, this is given by the following equations:
\begin{gather}
    \mPhi_{k,i}  = \phi  \Big ( \cF(\res{\lb}{n-1})_{k,i} \Big ),\quad
    \mA_{k,i} = \Big | \cF(\res{\lb}{n-1})_{k,i} \Big |, \nonumber \\
    \kappa_i^{(1)}, \ldots, \kappa_i^{(K)}  = \argtopk_{k \in \{2, \ldots, F\}} \nobreakspace \Big \{ \mA_{k,i} \Big \}, \nonumber \\
    \season{j,i}{n} = \sum_{k=1}^K  \mA_{\kappa_i^{(k)},i} \Big [ \cos (2 \pi f_{\kappa_i^{(k)}} j +  \mPhi_{\kappa_i^{(k)},i}) + \cos(2 \pi \bar{f}_{\kappa_i^{(k)}} j + \bar{\mPhi}_{\kappa_i^{(k)},i}) \Big ], \label{eq:seasonal-extrapolate}
\end{gather}
where \(\mPhi_{k,i}, \mA_{k,i}\) are the phase/amplitude of the \(k\)-th frequency for the \(i\)-th dimension, \(\argtopk\) returns the arguments of the top \(K\) amplitudes, \(K\) is a hyperparameter, \(f_k\) is the Fourier frequency of the corresponding index, and \(\bar{f}_k, \bar{\mPhi}_{k,i}\) are the Fourier frequency/amplitude of the corresponding conjugates. 

Finally, the latent seasonal representation of the $i$-th dimension for the lookback window is formulated as \(\season{\lb, i}{n} = [\season{t-L,i}{n}, \ldots, \season{t-1,i}{n}]\).
For the forecast horizon, the FA module extrapolates beyond the lookback window via, \(\season{\hrz, i}{n} = [\season{t,i}{n}, \ldots, \season{t+H-1,i}{n}]\).
Since \(K\) is a hyperparameter typically chosen for small values, the complexity for the FA mechanism is similarly \(\cO(L \log L)\).
\section{Experiments}
\label{sec:experiments}
This section presents extensive empirical evaluations on the LSTF task over 6 real world datasets, ETT, ECL, Exchange, Traffic, Weather, and ILI, coming from a variety of application areas (details in \cref{app:data}) for both multivariate and univariate settings. 
This is followed by an ablation study of the various contributing components, and interpretability experiments of our proposed model. An additional analysis on computational efficiency can be found in \cref{app:efficiency} for space.
For the main benchmark, datasets are split into train, validation, and test sets chronologically, following a 60/20/20 split for the ETT datasets and 70/10/20 split for other datasets. Inputs are zero-mean normalized and we use MSE and MAE as evaluation metrics. Further details on implementation and hyperparameters can be found in \cref{app:implementation}.

\subsection{Results}
\begin{table}[t]
\caption{Multivariate forecasting results over various forecast horizons. Best results are \textbf{bolded}, and second best results are \underline{underlined}.}
\label{tab:multivar-results}
\centering
\resizebox{\textwidth}{!}{
\begin{tabular}{c|c|cccccccccccccc}
\toprule
\multicolumn{2}{c}{Methods} & \multicolumn{2}{c}{{\shortname}} & \multicolumn{2}{c}{Autoformer} & \multicolumn{2}{c}{Informer} & \multicolumn{2}{c}{LogTrans} & \multicolumn{2}{c}{Reformer} & \multicolumn{2}{c}{LSTnet} & \multicolumn{2}{c}{LSTM} \\
\midrule
\multicolumn{2}{c}{Metrics} & MSE   & MAE   & MSE   & MAE   & MSE   & MAE   & MSE   & MAE   & MSE   & MAE   & MSE   & MAE   & MSE   & MAE \\
\midrule
\multirow{4}[2]{*}{\begin{sideways}ETTm2\end{sideways}} & 96    & \textbf{0.189} & \textbf{0.280} & \underline{0.255} & \underline{0.339} & 0.365 & 0.453 & 0.768 & 0.642 & 0.658 & 0.619 & 3.142 & 1.365 & 2.041 & 1.073 \\
      & 192   & \textbf{0.253} & \textbf{0.319} & \underline{0.281} & \underline{0.340} & 0.533 & 0.563 & 0.989 & 0.757 & 1.078 & 0.827 & 3.154 & 1.369 & 2.249 & 1.112 \\
      & 336   & \textbf{0.314} & \textbf{0.357} & \underline{0.339} & \underline{0.372} & 1.363 & 0.887 & 1.334 & 0.872 & 1.549 & 0.972 & 3.160 & 1.369 & 2.568 & 1.238 \\
      & 720   & \textbf{0.414} & \textbf{0.413} & \underline{0.422} & \underline{0.419} & 3.379 & 1.388 & 3.048 & 1.328 & 2.631 & 1.242 & 3.171 & 1.368 & 2.720 & 1.287 \\
\midrule
\multirow{4}[2]{*}{\begin{sideways}ECL\end{sideways}} & 96    & \textbf{0.187} & \textbf{0.304} & \underline{0.201} & \underline{0.317} & 0.274 & 0.368 & 0.258 & 0.357 & 0.312 & 0.402 & 0.680 & 0.645 & 0.375 & 0.437 \\
      & 192   & \textbf{0.199} & \textbf{0.315} & \underline{0.222} & \underline{0.334} & 0.296 & 0.386 & 0.266 & 0.368 & 0.348 & 0.433 & 0.725 & 0.676 & 0.442 & 0.473 \\
      & 336   & \textbf{0.212} & \textbf{0.329} & \underline{0.231} & \underline{0.338} & 0.300 & 0.394 & 0.280 & 0.380 & 0.350 & 0.433 & 0.828 & 0.727 & 0.439 & 0.473 \\
      & 720   & \textbf{0.233} & \textbf{0.345} & \underline{0.254} & \underline{0.361} & 0.373 & 0.439 & 0.283 & 0.376 & 0.340 & 0.420 & 0.957 & 0.811 & 0.980 & 0.814 \\
\midrule
\multirow{4}[2]{*}{\begin{sideways}Exchange\end{sideways}} & 96    & \textbf{0.085} & \textbf{0.204} & \underline{0.197} & \underline{0.323} & 0.847 & 0.752 & 0.968 & 0.812 & 1.065 & 0.829 & 1.551 & 1.058 & 1.453 & 1.049 \\
      & 192   & \textbf{0.182} & \textbf{0.303} & \underline{0.300} & \underline{0.369} & 1.204 & 0.895 & 1.040 & 0.851 & 1.188 & 0.906 & 1.477 & 1.028 & 1.846 & 1.179 \\
      & 336   & \textbf{0.348} & \textbf{0.428} & \underline{0.509} & \underline{0.524} & 1.672 & 1.036 & 1.659 & 1.081 & 1.357 & 0.976 & 1.507 & 1.031 & 2.136 & 1.231 \\
      & 720   & \textbf{1.025} & \textbf{0.774} & \underline{1.447} & \underline{0.941} & 2.478 & 1.310 & 1.941 & 1.127 & 1.510 & 1.016 & 2.285 & 1.243 & 2.984 & 1.427 \\
\midrule
\multirow{4}[2]{*}{\begin{sideways}Traffic\end{sideways}} & 96    & \textbf{0.607} & 0.392 & \underline{0.613} & \underline{0.388} & 0.719 & 0.391 & 0.684 & \textbf{0.384} & 0.732 & 0.423 & 1.107 & 0.685 & 0.843 & 0.453 \\
      & 192   & \underline{0.621} & 0.399 & \textbf{0.616} & \underline{0.382} & 0.696 & \textbf{0.379} & 0.685 & 0.390 & 0.733 & 0.420 & 1.157 & 0.706 & 0.847 & 0.453 \\
      & 336   & \underline{0.622} & \underline{0.396} & \textbf{0.622} & \textbf{0.337} & 0.777 & 0.420 & 0.733 & 0.408 & 0.742 & 0.420 & 1.216 & 0.730 & 0.853 & 0.455 \\
      & 720   & \textbf{0.632} & \underline{0.396} & \underline{0.660} & 0.408 & 0.864 & 0.472 & 0.717 & \textbf{0.396} & 0.755 & 0.423 & 1.481 & 0.805 & 1.500 & 0.805 \\
\midrule
\multirow{4}[2]{*}{\begin{sideways}Weather\end{sideways}} & 96    & \textbf{0.197} & \textbf{0.281} & \underline{0.266} & \underline{0.336} & 0.300 & 0.384 & 0.458 & 0.490 & 0.689 & 0.596 & 0.594 & 0.587 & 0.369 & 0.406 \\
      & 192   & \textbf{0.237} & \textbf{0.312} & \underline{0.307} & \underline{0.367} & 0.598 & 0.544 & 0.658 & 0.589 & 0.752 & 0.638 & 0.560 & 0.565 & 0.416 & 0.435 \\
      & 336   & \textbf{0.298} & \textbf{0.353} & \underline{0.359} & \underline{0.359} & 0.578 & 0.523 & 0.797 & 0.652 & 0.639 & 0.596 & 0.597 & 0.587 & 0.455 & 0.454 \\
      & 720   & \textbf{0.352} & \textbf{0.388} & \underline{0.419} & \underline{0.419} & 1.059 & 0.741 & 0.869 & 0.675 & 1.130 & 0.792 & 0.618 & 0.599 & 0.535 & 0.520 \\
\midrule
\multirow{4}[2]{*}{\begin{sideways}ILI\end{sideways}} & 24    & \textbf{2.527} & \textbf{1.020} & \underline{3.483} & \underline{1.287} & 5.764 & 1.677 & 4.480 & 1.444 & 4.400 & 1.382 & 6.026 & 1.770 & 5.914 & 1.734 \\
      & 36    & \textbf{2.615} & \textbf{1.007} & \underline{3.103} & \underline{1.148} & 4.755 & 1.467 & 4.799 & 1.467 & 4.783 & 1.448 & 5.340 & 1.668 & 6.631 & 1.845 \\
      & 48    & \textbf{2.359} & \textbf{0.972} & \underline{2.669} & \underline{1.085} & 4.763 & 1.469 & 4.800 & 1.468 & 4.832 & 1.465 & 6.080 & 1.787 & 6.736 & 1.857 \\
      & 60    & \textbf{2.487} & \textbf{1.016} & \underline{2.770} & \underline{1.125} & 5.264 & 1.564 & 5.278 & 1.560 & 4.882 & 1.483 & 5.548 & 1.720 & 6.870 & 1.879 \\
\bottomrule
\end{tabular}%
}
\end{table}

For the multivariate benchmark, baselines include recently proposed time-series/efficient Transformers -- Autoformer, Informer, LogTrans, and Reformer \cite{Kitaev2020Reformer}, and RNN variants -- LSTnet \cite{lai2018modeling}, and LSTM \cite{hochreiter1997long}. Univariate baselines further include N-BEATS \cite{oreshkin2019n}, DeepAR \cite{salinas2020deepar}, ARIMA, Prophet \cite{taylor2018forecasting}, and AutoETS \cite{bhatnagar2021merlion}. We obtain baseline results from the following papers: \cite{wu2021autoformer, zhou2021informer}, and further run AutoETS from the Merlion library \cite{bhatnagar2021merlion}.
\cref{tab:multivar-results} summarize the results of {\shortname} against top performing baselines on a selection of datasets, for the multivariate setting, and \cref{tab:univar-results} in \cref{app:univar-results} for space. Results for {\shortname} are averaged over three runs (standard deviation in \cref{app:sd}).

Overall, {\shortname} achieves state-of-the-art performance, achieving the best performance (across all datasets/settings, based on MSE) on 35 out of 40 settings for the multivariate case, and 17 out of 23 for the univariate case. Notably, on Exchange, a dataset with no obvious periodic patterns, {\shortname} demonstrates an average (over forecast horizons) improvemnt of 39.8\% over the best performing baseline, evidencing its strong trend forecasting capabilities. We highlight that for cases where {\shortname} does not achieve the best performance, it is still highly competitive, and is always within the top 2 performing methods, based on MSE, for 40 out of 40 settings in the multivariate benchmark , and 21 out of 23 settings of the univariate case.

\subsection{Ablation Study}
\begin{table}[h]
\caption{Ablation study on the various components of ETSformer, on the horizon\(=24\) setting.}
\label{tab:component-ablation}
\centering
\scalebox{0.65}{
\begin{tabular}{c|ccccc}
\toprule
\multicolumn{2}{c}{Datasets} & ETTh2 & ETTm2 & ECL   & Traffic \\
\midrule
\multirow{2}[2]{*}{ETSformer} & MSE   & \textbf{0.262} & \textbf{0.110} & \textbf{0.163} & \textbf{0.571} \\
      & MAE   & \textbf{0.337} & \textbf{0.222} & \textbf{0.287} & \textbf{0.373} \\
\midrule
\multirow{2}[2]{*}{w/o Level} & MSE   & 0.434 & 0.464 & 0.275 & 0.649 \\
      & MAE   & 0.466 & 0.518 & 0.373 & 0.393 \\
\midrule
\multirow{2}[2]{*}{w/o Season} & MSE   & 0.521 & 0.131 & 0.696 & 1.334 \\
      & MAE   & 0.450 & 0.236 & 0.677 & 0.779 \\
\midrule
\multirow{2}[2]{*}{w/o Growth} & MSE   & 0.290 & 0.115 & 0.167 & 0.583 \\
      & MAE   & 0.359 & 0.226 & 0.288 & 0.383 \\
\midrule
\multirow{2}[2]{*}{MH-ESA \(\to\) MHA} & MSE   & 0.656 & 0.343 & 0.205 & 0.586 \\
      & MAE   & 0.639 & 0.451 & 0.323 & 0.380 \\
\bottomrule
\end{tabular}%
}
\end{table}

We study the contribution of each major component which the final forecast is composed of level, growth, and seasonality. 
\cref{tab:component-ablation} first presents the performance of the full model, and subsequently, the performance of the resulting model by removing each component.
We observe that the composition of level, growth, and season provides the most accurate forecasts across a variety of application areas, and removing any one component results in a deterioration. In particular, estimation of the level of the time-series is critical. We also analyse the case where MH-ESA is replaced with a vanilla multi-head attention, and observe that our trend attention formulation indeed is more effective.

\begin{figure}[t]
\vspace{-0.1in}
    \centering
    \begin{subfigure}[b]{\columnwidth}
    \centering
    \includegraphics[width=0.49\textwidth]{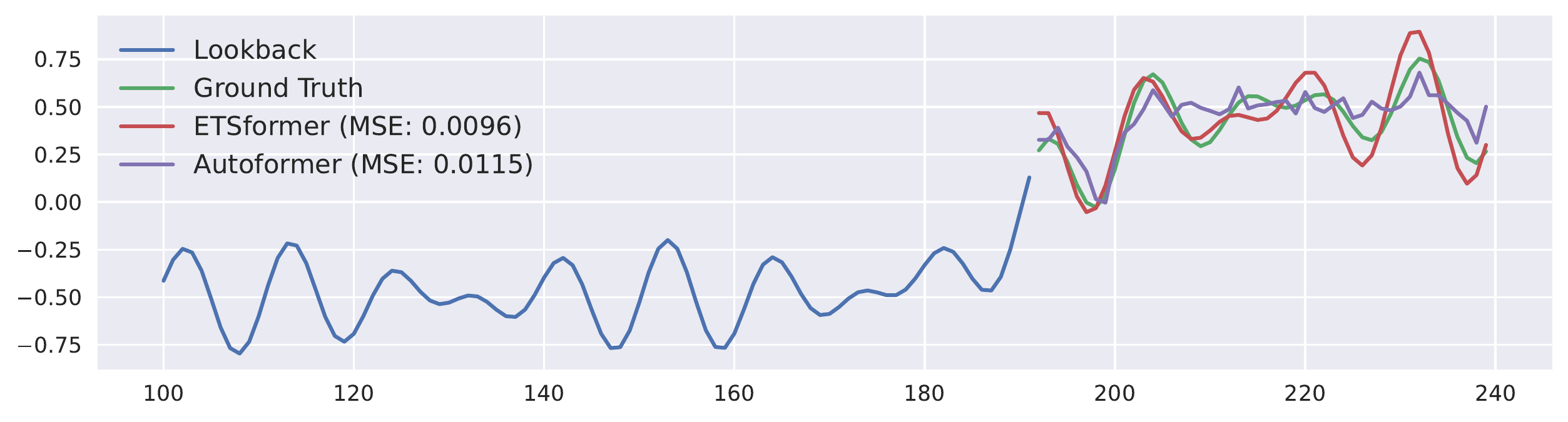}
    \includegraphics[width=0.49\textwidth]{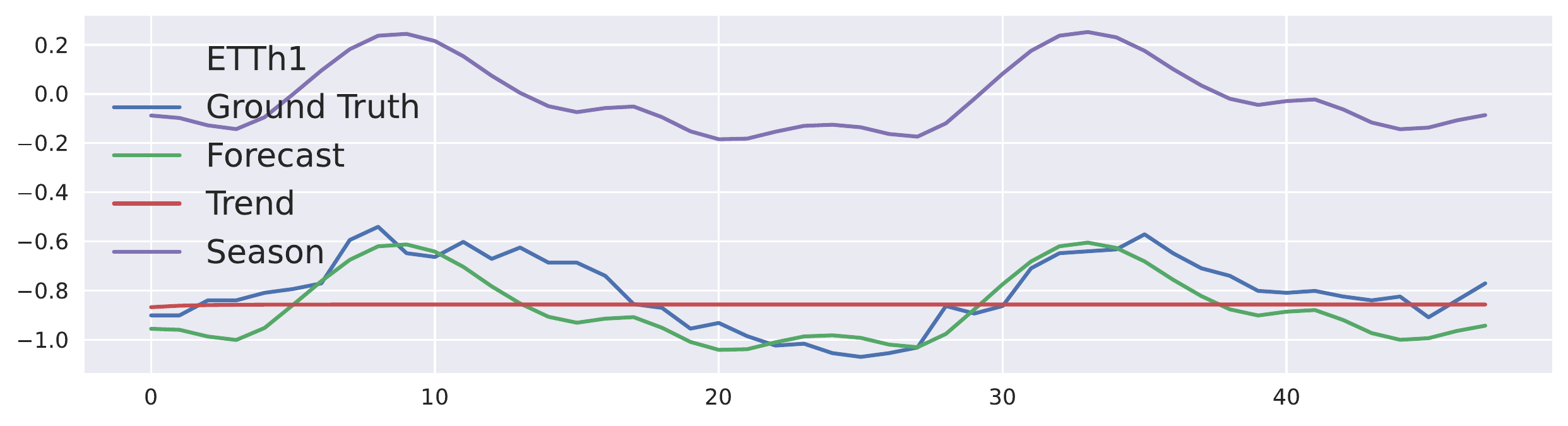}
    \end{subfigure}
    \begin{subfigure}[b]{\columnwidth}
    \centering
    \includegraphics[width=0.49\textwidth]{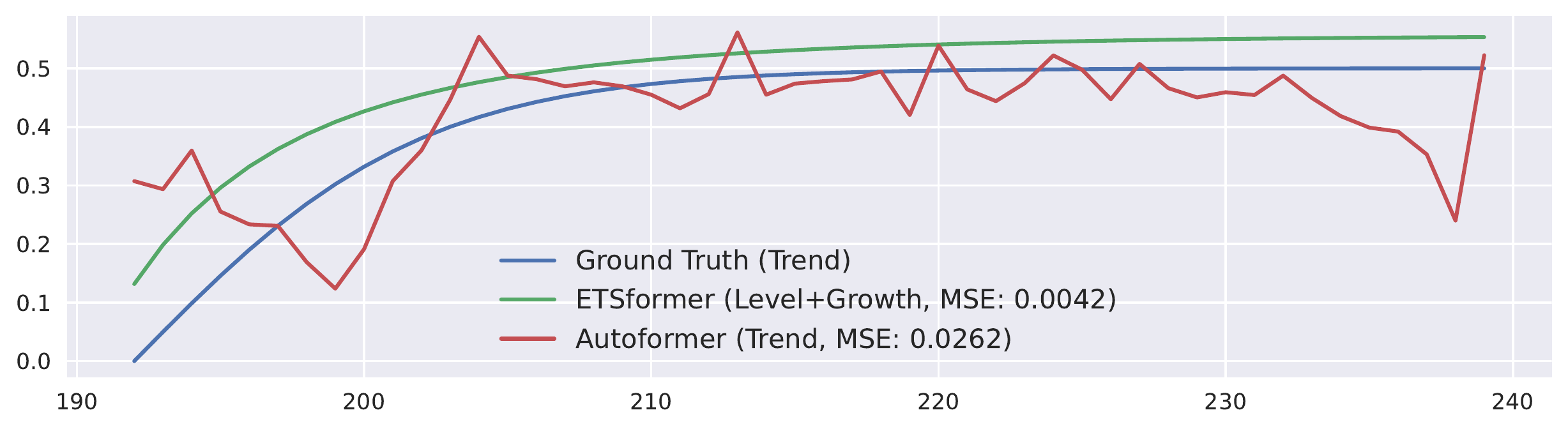}
    \includegraphics[width=0.49\textwidth]{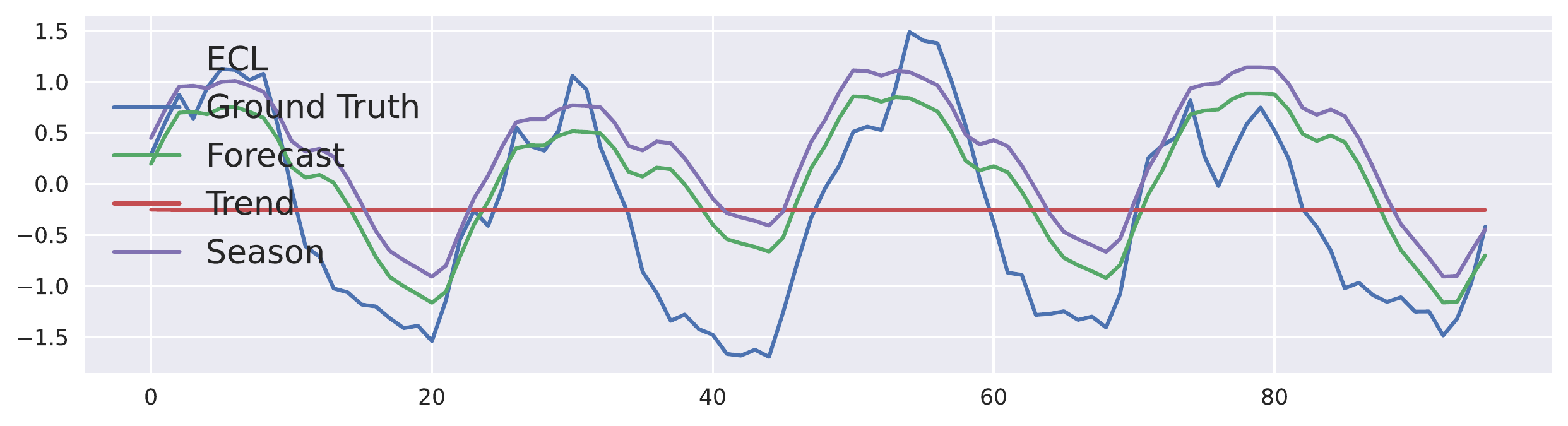}
    \end{subfigure}
    \begin{subfigure}[b]{\columnwidth}
    \centering
    \includegraphics[width=0.49\textwidth]{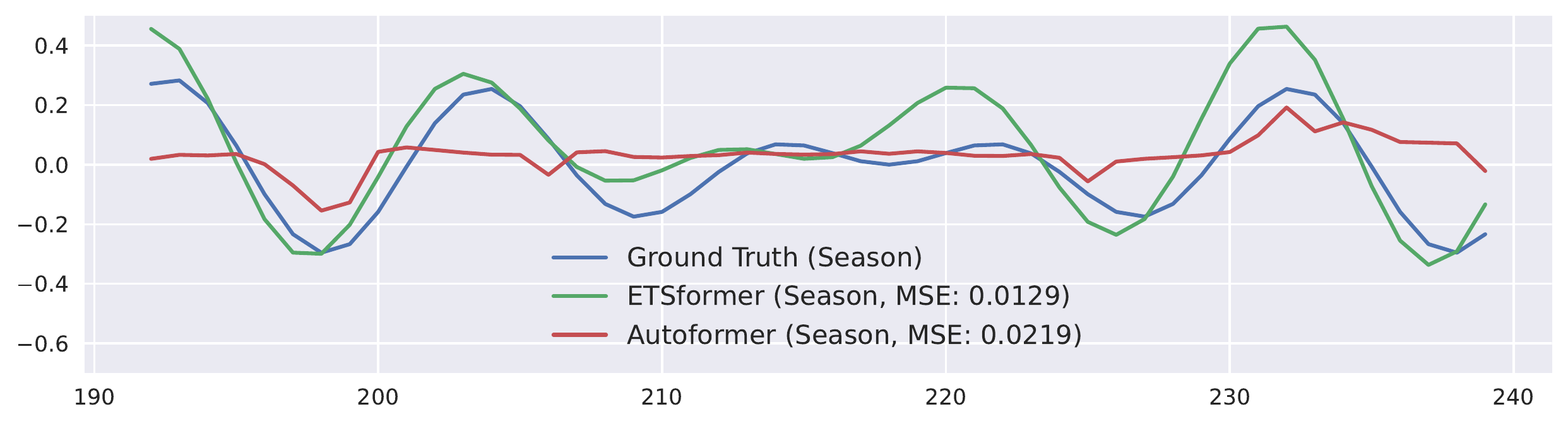}
    \includegraphics[width=0.49\textwidth]{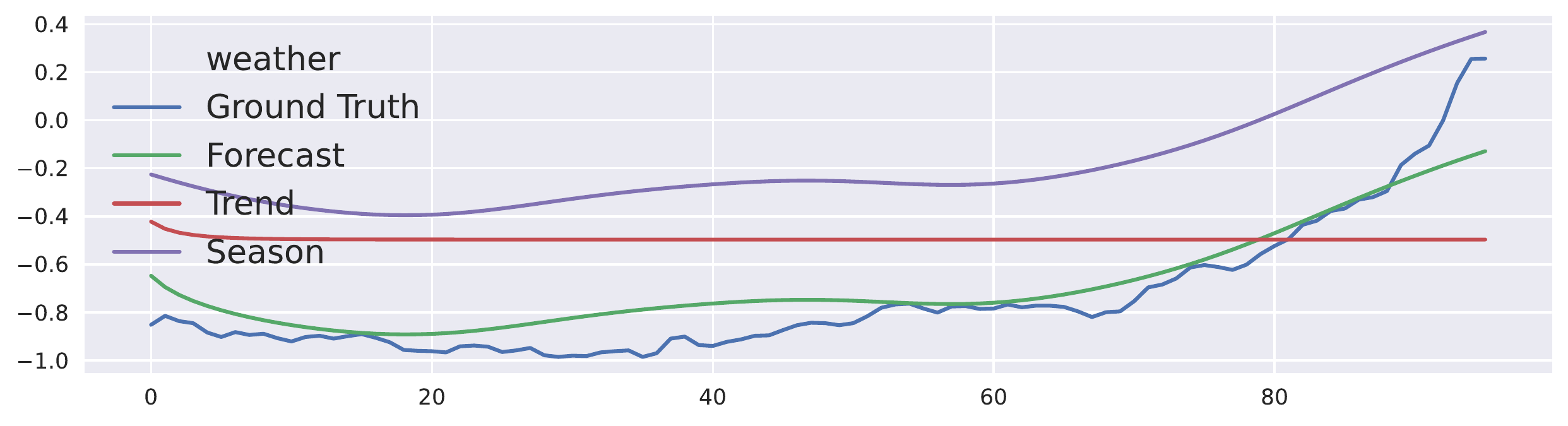}
    \end{subfigure}
    \caption{
    Left: Visualization of decomposed forecasts from ETSformer and Autoformer on a synthetic datset. 
    (i) Ground truth and non-decomposed forecasts of ETSformer and Autoformer on synthetic data.
    (ii) Trend component.
    (iii) Seasonal component.
    The data sample on which Autoformer obtained lowest MSE was selected for visualization.
    Right: Visualization of decomposed forecasts from ETSformer on real world datasets, ETTh1, ECL, and Weather.
    Note that season is zero-centered, and trend successfully tracks the level of the time-series. Due to the long sequence forecasting setting and with a damping, the growth component is not visually obvious, but notice for the Weather dataset, the trend pattern is has a strong downward slope initially (near time step 0), and is quickly damped.
    }
    \label{fig:interpretability}
\end{figure}
\subsection{Interpretability}
{\shortname} generates forecasts based on a composition of interpretable time-series components. This means we can visualize each component individually, and understand how seasonality and trend affects the forecasts. We showcase this ability in \cref{fig:interpretability} on both synthetic and real world data. Experiments with synthetic data are crucial in this case, since we are not able to obtain the ground truth decomposition from real world data. {\shortname} is first trained on the synthetic dataset (details in \cref{app:synthetic}) with clear (nonlinear) trend and seasonality patterns which we can control. Given a lookback window (without noise), we visualize the forecast, as well as decomposed trend and seasonal forecasts.
{\shortname} successfully forecasts interpretable level, trend (level + growth), and seasonal components, as observed in the trend and seasonality components closely tracking the ground truth patterns. Despite obtaining a low MSE, the competing decomposition based approach, Autoformer, struggles to disambiguate between trend and seasonality. 
\vspace{-0.1in}
\section{Conclusion}
\label{sec:conclusion}
Inspired by the classical exponential smoothing methods and emerging Transformer approaches for time-series forecasting, we proposed {\shortname}, a novel Transformer-based architecture for time-series forecasting which learns level, growth, and seasonal latent representations and their complex dependencies. {\shortname} leverages the novel Exponential Smoothing Attention and Frequency Attention mechanisms which are more effective at modeling time-series than vanilla self-attention mechanism, and at the same time achieves \(\cO(L \log L)\) complexity, where \(L\) is the length of lookback window. 
Our extensive empirical evaluation shows that {\shortname} achieves state-of-the-art performance, beating competing baselines in 35 out of 40 and 17 out of 23 settings for multivariate and univariate forecasting respectively.
Future directions include including additional covariates such as holiday indicators and other dummy variables to consider holiday effects which cannot be captured by the FA mechanism.

\newpage
{\small
\bibliography{neurips_2022}}
\bibliographystyle{plainnat}


\newpage
\appendix
\section{Exponential Smoothing Attention}
\label{app:esa-algorithm}

\subsection{Exponential Smoothing Attention Matrix}
\label{subapp:esa-matrix}
\begin{gather*}
\mA_{\mathrm{ES}} = \begin{bmatrix}
    (1-\alpha)^1 & \alpha  & 0 & 0 & \ldots & 0 \\
    (1-\alpha)^2 & \alpha (1-\alpha) & \alpha & 0 & \ldots & 0 \\
    (1-\alpha)^3 & \alpha (1-\alpha)^2 & \alpha (1-\alpha) & \alpha & \ldots & 0 \\
    \vdots & \vdots & \vdots & \vdots & \ddots & \vdots \\
    (1 - \alpha)^L & \alpha (1-\alpha)^{L-1} & \ldots & \alpha (1-\alpha)^j & \ldots & \alpha
\end{bmatrix}
\end{gather*}

\subsection{Efficient Exponential Smoothing Attention Algorithm}
\label{subapp:esa-algo}
\begin{algorithm}[H]
   \caption{PyTorch-style pseudocode of efficient \(\ESA\)}
   \label{alg:efficient-esa}
   \fontsize{8pt}{0em}\selectfont \texttt{conv1d\_fft}: efficient convolution operation implemented with fast Fourier transform (\cref{app:esa-algorithm}, \cref{alg:conv-fft}), \\ \texttt{outer}: outer product
\begin{lstlisting}[language=python]
# V: value matrix, shape: L x d
# v0: initial state, shape: d
# alpha: smoothing parameter, shape: 1

# obtain exponentially decaying weights 
# and compute weighted combination
powers = arange(L)  # L
weight = alpha * (1 - alpha) ** flip(powers)  # L
output = conv1d_fft(V, weight, dim=0)  # L x d

# compute contribution from initial state
init_weight = (1 - alpha) ** (powers + 1)  # L
init_output = outer(init_weight, v0)  # L x d
return init_output + output
\end{lstlisting}
\end{algorithm}

\subsection{Level Smoothing via Exponential Smoothing Attention}
\label{subapp:level-esa}
\begin{align*}
     \level{t}{n} & = \valpha * ( \level{t}{n-1} - \season{t}{n} ) + (1 - \valpha) * ( \level{t-1}{n} + \trend{t-1}{n} ) \\
     & =  \valpha * (\level{t}{n-1} - \season{t}{n}) + (1 - \valpha) * \trend{t-1}{n} \\
     & \hspace{3em} + (1 - \valpha) * [\valpha * (\level{t-1}{n-1} - \season{t-1}{n}) + (1 - \valpha) * (\level{t-2}{n} + \trend{t-2}{n}) ] \\
     & = \valpha * (\level{t}{n-1} - \season{t}{n}) + \valpha * (1 - \valpha) * (\level{t-1}{n-1} - \season{t-1}{n}) \\
     & \hspace{3em} + (1 - \valpha) * \trend{t-1}{n} + (1 - \valpha )^2 * \trend{t-2}{n} \\
     & \hspace{3em} + (1 - \valpha)^2 [\valpha * (\level{t-2}{n-1} - \season{t-2}{n}) + (1 - \valpha) * (\level{t-3}{n} + \trend{t-3}{n})] \\
     & \vdotswithin{=} \\
     & = (1 - \valpha)^t (\level{0}{n} - \season{0}{n}) + \sum_{j=0}^{t-1} \valpha * (1 - \valpha)^j * ( \level{t-j}{n-1} - \season{t-j}{n}) + \sum_{k=1}^t (1 - \valpha)^k * \trend{t-k}{n} \\
     & = \ESA(\level{\lb}{n-1} - \season{\lb}{n}) + \sum_{k=1}^t (1 - \valpha)^k * \trend{t-k}{n}
\end{align*}
Based on the above expansion of the level equation, we observe that \(\level{n}{t}\) can be computed by a sum of two terms, the first of which is given by an \(\ESA\) term, and we finally, we note that the second term can also be calculated using the conv1d\_fft algorithm, resulting in a fast implementation of level smoothing.

\newpage
\subsection{Further Details on ESA Implementation}
\label{subapp:esa-implementation}
\begin{minipage}[t]{0.49\textwidth}
\begin{algorithm}[H]\small
  \caption{PyTorch-style pseudocode of naive \(\ESA\)}
  \label{alg:naive-esa}
  \fontsize{8pt}{0em}\selectfont \texttt{mm}: matrix multiplication, \texttt{outer}: outer product \\ 
  \texttt{repeat}: einops style tensor operations, \\
  \texttt{gather}: gathers values along an axis specified by dim
\begin{lstlisting}[language=python]
# V: value matrix, shape: L x d
# v0: initial state, shape: d
# alpha: smoothing parameter, shape: 1

L, d = V.shape

# obtain exponentially decaying weights
powers = arange(L)  # L
weight = alpha * (1 - alpha).pow(flip(powers))  # L

# perform a strided roll operation
# rolls a matrix along the columns in a strided manner
# i.e. first row is shifted right by L-1 positions, 
# second row is shifted L-2, ..., last row is shifted by 0.
weight = repeat(weight, 'L -> T L', T=L)  # L x L
indices = repeat(arange(L), 'L -> T L', T=L)
indices = (indices - (arange(L) + 1).unsqueeze(1)) % L
weight = gather(weight, dim=-1, index=indices)

# triangle masking to achieve the exponential smoothing attention matrix
weight = triangle_causal_mask(weight)

output = mm(weight, V)

init_weight = (1 - alpha) ** (powers + 1)
init_output = outer(init_weight, v0)

return init_output + output
\end{lstlisting}
\end{algorithm}
\end{minipage}
\begin{minipage}[t]{0.49\textwidth}
\begin{algorithm}[H]
  \caption{PyTorch-style pseudocode of conv1d\_fft}
  \label{alg:conv-fft}
  \fontsize{8pt}{0em}\selectfont 
  \texttt{next_fast_len}: find the next fast size of input data to fft, for zero-padding, etc. \\
  \texttt{rfft}: compute the one-dimensional discrete Fourier Transform for real input \\
  \texttt{x.conj()}: return the complex conjugate, element-wise \\ 
  \texttt{irfft}: computes the inverse of rfft \\
  \texttt{roll}: roll array elements along a given axis \\
  \texttt{index\_select}: returns a new tensor which index es the input tensor along dimension dim using the entries in index
\begin{lstlisting}[language=python]
# V: value matrix, shape: L x d
# weight: exponential smoothing attention vector, shape: L
# dim: dimension to perform convolution on

# obtain lengths of sequence to perform convolution on
N = V.size(dim)
M = weight.size(dim)

# Fourier transform on inputs
fast_len = next_fast_len(N + M - 1)
F_V = rfft(V, fast_len, dim=dim)
F_weight = rfft(weight, fast_len, dim=dim)

# multiplication and inverse
F_V_weight = F_V * F_weight.conj()
out = irfft(F_V_weight, fast_len, dim=dim)
out = out.roll(-1, dim=dim)

# select the correct indices
idx = range(fast_len - N, fast_len)
out = out.index_select(dim, idx)

return out
\end{lstlisting}
\end{algorithm}
\end{minipage}

\cref{alg:naive-esa} describes the naive implementation for ESA by first constructing the exponential smoothing attention matrix, \(\mA_{\mathrm{ES}}\), and performing the full matrix-vector multiplication. 
Efficient \(\ESA\) relies on \cref{alg:conv-fft}, to achieve an \(\cO(L \log L)\) complexity, by speeding up the matrix-vector multiplication. Due to the structure lower triangular structure of \(\mA_{\mathrm{ES}}\) (ignoring the first column), we note that performing a matrix-vector multiplication with it is equivalent to performing a convolution with the last row. \cref{alg:conv-fft} describes the pseudocode for fast convolutions using fast Fourier transforms.

\section{Discrete Fourier Transform}
\label{app:dft}
The DFT of a sequence with regular intervals, \(\vx = (\evx_0, \evx_1, \ldots, \evx_{N-1})\) is a sequence of complex numbers,
\[ \evc_k = \sum_{n=0}^{N-1} \evx_n \cdot \exp (- i 2 \pi k n / N),\]
for \(k=0, 1, \ldots, N-1\), where \(c_k\) are known as the Fourier coefficients of their respective Fourier frequencies. 
Due to the conjugate symmetry of DFT for real-valued signals, we simply consider the first \(\lfloor N/2 \rfloor + 1\) Fourier coefficients and thus we denote the DFT as \(\cF: \R^N \to \sC^{\lfloor N/2 \rfloor +1}\).
The DFT maps a signal to the frequency domain, where each Fourier coefficient can be uniquely represented by the amplitude, \(|\evc_k|\), and the phase, \(\phi(\evc_k)\),
\begin{align*}
    |\evc_k| & = \sqrt{\mathfrak{R}\{\evc_k\}^2 + \mathfrak{I}\{\evc_k\}^2}
    & 
    \phi(\evc_k) & = \tan^{-1} \bigg( \frac{\mathfrak{I}\{\evc_k\}}{\mathfrak{R}\{\evc_k\}} \bigg)
\end{align*}
where \(\mathfrak{R}\{\evc_k\}\) and \(\mathfrak{I}\{\evc_k\}\) are the real and imaginary components of \(\evc_k\) respectively.
Finally, the inverse DFT maps the frequency domain representation back to the time domain,
\[\evx_n = \cF^{-1}(\vc)_n = \frac{1}{N} \sum_{k=0}^{N-1} c_k \cdot \exp(i 2 \pi k n / N),\]

\section{Implementation Details}
\label{app:implementation}
\subsection{Hyperparameters}
For all experiments, we use the same hyperparameters for the encoder layers, decoder stacks, model dimensions, feedforward layer dimensions, number of heads in multi-head exponential smoothing attention, and kernel size for input embedding as listed in \cref{tab:hyperparams}. We perform hyperparameter tuning via a grid search over the number of frequencies \(K\), lookback window size, and learning rate, selecting the settings which perform the best on the validation set based on MSE (on results averaged over three runs). The search range is reported in \cref{tab:hyperparams}, where the lookback window size search range was decided to be set as the values for the horizon sizes for the respective datasets.
\begin{table}[H]
  \centering
  \caption{Hyperparameters used in {\shortname}.}
\begin{tabular}{ll}
\toprule
Hyperparameter & Value \\
\midrule
Encoder layers & 2 \\
Decoder stacks & 2 \\
Model dimension & 512 \\
Feedforward dimension & 2048 \\
Multi-head ESA heads & 8 \\
Input embedding kernel size & 3 \\
K     & \(K \in \{0,1,2,3\}\) \\
Lookback window size & \(L \in \{96, 192, 336, 720\}\) \\
Lookback window size (ILI) & \(L \in \{24, 36, 48, 60\}\) \\
Learning rate & \(lr \in \{1\mathrm{e}{-3}, 3\mathrm{e}{-4}, 1\mathrm{e}{-4}, 3\mathrm{e}{-5}, 1\mathrm{e}{-5}\}\) \\
\bottomrule
\end{tabular}%
  \label{tab:hyperparams}%
\end{table}%

\subsection{Optimization}
We use the Adam optimizer \cite{kingma2015adam} with \(\beta_1 = 0.9\), \(\beta_2 = 0.999\), and \(\epsilon = 1e-08\), and a batch size of 32.
We schedule the learning rate with linear warmup over 3 epochs, and cosine annealing thereafter for a total of 15 training epochs for all datasets. The minimum learning rate is set to 1e-30. For smoothing and damping parameters, we set the learning rate to be 100 times larger and do not use learning rate scheduling. Training was done on an Nvidia A100 GPU.

\subsection{Regularization}
We apply two forms of regularization during the training phase.

\paragraph{Data Augmentations}
We utilize a composition of three data augmentations, applied in the following order - scale, shift, and jitter, activating with a probability of 0.5.

\begin{enumerate}
    \item Scale -- The time-series is scaled by a single random scalar value, obtained by sampling $\epsilon \sim \mathcal{N}(0, 0.2)$, and each time step is $\tilde{x}_t = \epsilon x_t$.
    \item Shift -- The time-series is shifted by a single random scalar value, obtained by sampling $\epsilon \sim \mathcal{N}(0, 0.2)$ and each time step is $\tilde{x}_t = x_t + \epsilon$.
    \item Jitter -- I.I.D. Gaussian noise is added to each time step, from a distribution $\epsilon_t \sim \mathcal{N}(0, 0.2)$, where each time step is now $\tilde{x}_t = x_t + \epsilon_t$.
\end{enumerate}

\paragraph{Dropout}
We apply dropout \cite{srivastava2014dropout} with a rate of \(p=0.2\) across the model. Dropout is applied on the outputs of the Input Embedding, Frequency Self-Attention and Multi-Head ES Attention blocks, in the Feedforward block (after activation and before normalization), on the attention weights, as well as damping weights.

\section{Datasets}
\label{app:data}

\textbf{ETT}\footnote{\url{https://github.com/zhouhaoyi/ETDataset}} 
Electricity Transformer Temperature \cite{zhou2021informer} is a multivariate time-series dataset, comprising of load and oil temperature data recorded every 15 minutes from electricity transformers. ETT consists of two variants, ETTm and ETTh, whereby ETTh is the hourly-aggregated version of ETTm, the original 15 minute level dataset.

\textbf{ECL}\footnote{\url{lhttps://archive.ics.uci.edu/ml/datasets/ElectricityLoadDiagrams20112014}}
Electricity Consuming Load measures the electricity consumption of 321 households clients over two years, the original dataset was collected at the 15 minute level, but is pre-processed into an hourly level dataset.

\textbf{Exchange}\footnote{\url{https://github.com/laiguokun/multivariate-time-series-data}}
Exchange \cite{lai2018modeling} tracks the daily exchange rates of eight countries (Australia, United Kingdom, Canada, Switzerland, China, Japan, New Zealand, and Singapore) from 1990 to 2016.

\textbf{Traffic}\footnote{\url{https://pems.dot.ca.gov/}}
Traffic is an hourly dataset from the California Department of Transportation describing road occupancy rates in San Francisco Bay area freeways.

\textbf{Weather}\footnote{\url{https://www.bgc-jena.mpg.de/wetter/}}
Weather measures 21 meteorological indicators like air temperature, humidity, etc., every 10 minutes for the year of 2020.

\textbf{ILI}\footnote{\url{https://gis.cdc.gov/grasp/fluview/fluportaldashboard.html}}
Influenza-like Illness records the ratio of patients seen with ILI and the total number of patients on a weekly basis, obtained by the Centers for Disease Control and Prevention of the United States between 2002 and 2021.

\section{Synthetic Dataset}
\label{app:synthetic}
The synthetic dataset is constructed by a combination of trend and seasonal component. 
Each instance in the dataset has a lookack window length of 192 and forecast horizon length of 48. 
The trend pattern follows a nonlinear, saturating pattern, \(b(t) = \frac{1}{1 + \exp{\beta_0(t-\beta_1)}}\), where \(\beta_0=-0.2, \beta_1=192\).
The seasonal pattern follows a complex periodic pattern formed by a sum of sinusoids. Concretely, \(s(t) = A_1 \cos(2 \pi f_1 t) + A_2 \cos(2 \pi f_2 t\), where \(f_1 = 1/10, f_2 = 1/13\) are the frequencies, \(A_1 = A_2 = 0.15\) are the amplitudes.
During training phase, we use an additional noise component by adding i.i.d. gaussian noise with 0.05 standard deviation.
Finally, the \(i\)-th instance of the dataset is \(x_i = [x_i(1), x_i(2), \ldots, x_i(192+48)]\), where \(x_i(t) = b(t) + s(t + i) + \epsilon\).

\newpage
\section{Univariate Forecasting Benchmark}
\label{app:univar-results}
\begin{table}[h]
\caption{Univariate forecasting results over various forecast horizons. Best results are \textbf{bolded}, and second best results are \underline{underlined}.}
\label{tab:univar-results}
\centering
\resizebox{\textwidth}{!}{
\begin{tabular}{c|c|cccccccccccccccc}
\toprule
\multicolumn{2}{c}{Methods} & \multicolumn{2}{c}{{\shortname}} & \multicolumn{2}{c}{Autoformer} & \multicolumn{2}{c}{Informer} & \multicolumn{2}{c}{N-BEATS} & \multicolumn{2}{c}{DeepAR} & \multicolumn{2}{c}{Prophet} & \multicolumn{2}{c}{ARIMA} & \multicolumn{2}{c}{AutoETS} \\
\midrule
\multicolumn{2}{c}{Metrics} & MSE   & MAE   & MSE   & MAE   & MSE   & MAE   & MSE   & MAE   & MSE   & MAE   & MSE   & MAE   & MSE   & MAE   & MSE   & MAE \\
\midrule
\multirow{4}[2]{*}{\begin{sideways}ETTm2\end{sideways}} & 96    & \underline{0.080} & \underline{0.212} & \textbf{0.065} & \textbf{0.189} & 0.088 & 0.225 & 0.082 & 0.219 & 0.099 & 0.237 & 0.287 & 0.456 & 0.211 & 0.362 & 0.794 & 0.617 \\
      & 192   & 0.150 & 0.302 & \textbf{0.118} & \textbf{0.256} & 0.132 & 0.283 & \underline{0.120} & \underline{0.268} & 0.154 & 0.310 & 0.312 & 0.483 & 0.261 & 0.406 & 1.078 & 0.740 \\
      & 336   & \underline{0.175} & \underline{0.334} & \textbf{0.154} & \textbf{0.305} & 0.180 & 0.336 & 0.226 & 0.370 & 0.277 & 0.428 & 0.331 & 0.474 & 0.317 & 0.448 & 1.279 & 0.822 \\
      & 720   & 0.224 & 0.379 & \textbf{0.182} & \textbf{0.335} & 0.300 & 0.435 & \underline{0.188} & \underline{0.338} & 0.332 & 0.468 & 0.534 & 0.593 & 0.366 & 0.487 & 1.541 & 0.924 \\
\midrule
\multirow{4}[2]{*}{\begin{sideways}Exchange\end{sideways}} & 96    & \textbf{0.099} & \textbf{0.230} & 0.241 & 0.299 & 0.591 & 0.615 & 0.156 & 0.299 & 0.417 & 0.515 & 0.828 & 0.762 & \underline{0.112} & \underline{0.245} & 0.192 & 0.316 \\
      & 192   & \textbf{0.223} & \textbf{0.353} & \underline{0.273} & 0.665 & 1.183 & 0.912 & 0.669 & 0.665 & 0.813 & 0.735 & 0.909 & 0.974 & 0.304 & \underline{0.404} & 0.355 & 0.442 \\
      & 336   & \textbf{0.421} & \textbf{0.497} & \underline{0.508} & 0.605 & 1.367 & 0.984 & 0.611 & 0.605 & 1.331 & 0.962 & 1.304 & 0.988 & 0.736 & 0.598 & 0.577 & \underline{0.578} \\
      & 720   & 1.114 & \textbf{0.807} & \textbf{0.991} & \underline{0.860} & 1.872 & 1.072 & \underline{1.111} & \underline{0.860} & 1.890 & 1.181 & 3.238 & 1.566 & 1.871 & 0.935 & 1.242 & 0.865 \\
\bottomrule
\end{tabular}%
}
\end{table}

\section{{\shortname} Standard Deviation}
\label{app:sd}
\begin{table}[h]
    \caption{{\shortname} main benchmark results with standard deviation. Experiments are performed over three runs.}
    \begin{subtable}[t]{.5\linewidth}
      \centering
        \caption{Multivariate benchmark.}
\begin{tabular}{c|c|cc}
\toprule
\multicolumn{2}{c}{Metrics} & MSE (SD) & MAE (SD) \\
\midrule
\multirow{4}[2]{*}{\begin{sideways}ETTm2\end{sideways}} & 96    & 0.189 (0.002) & 0.280 (0.001) \\
      & 192   & 0.253 (0.002) & 0.319 (0.001) \\
      & 336   & 0.314 (0.001) & 0.357 (0.001) \\
      & 720   & 0.414 (0.000) & 0.413 (0.001) \\
\midrule
\multirow{4}[2]{*}{\begin{sideways}ECL\end{sideways}} & 96    & 0.187 (0.001) & 0.304 (0.001) \\
      & 192   & 0.199 (0.001) & 0.315 (0.002) \\
      & 336   & 0.212 (0.001) & 0.329 (0.002) \\
      & 720   & 0.233 (0.006) & 0.345 (0.006) \\
\midrule
\multirow{4}[2]{*}{\begin{sideways}Exchange\end{sideways}} & 96    & 0.085 (0.000) & 0.204 (0.001) \\
      & 192   & 0.182 (0.003) & 0.303 (0.002) \\
      & 336   & 0.348 (0.004) & 0.428 (0.003) \\
      & 720   & 1.025 (0.031) & 0.774 (0.014) \\
\midrule
\multirow{4}[2]{*}{\begin{sideways}Traffic\end{sideways}} & 96    & 0.607 (0.005) & 0.392 (0.005) \\
      & 192   & 0.621 (0.015) & 0.399 (0.013) \\
      & 336   & 0.622 (0.003) & 0.396 (0.003) \\
      & 720   & 0.632 (0.004) & 0.396 (0.004) \\
\midrule
\multirow{4}[2]{*}{\begin{sideways}Weather\end{sideways}} & 96    & 0.197 (0.007) & 0.281 (0.008) \\
      & 192   & 0.237 (0.005) & 0.312 (0.004) \\
      & 336   & 0.298 (0.003) & 0.353 (0.003) \\
      & 720   & 0.352 (0.007) & 0.388 (0.002) \\
\midrule
\multirow{4}[2]{*}{\begin{sideways}ILI\end{sideways}} & 24    & 2.527 (0.061) & 1.020 (0.021) \\
      & 36    & 2.615 (0.103) & 1.007 (0.013) \\
      & 48    & 2.359 (0.056) & 0.972 (0.011) \\
      & 60    & 2.487 (0.006) & 1.016 (0.007) \\
\bottomrule
\end{tabular}%
    \end{subtable}%
    \begin{subtable}[t]{.5\linewidth}
      \centering
        \caption{Univariate benchmark.}
\begin{tabular}{c|c|cc}
\toprule
\multicolumn{2}{c}{Metrics} & MSE (SD) & MAE (SD) \\
\midrule
\multirow{4}[2]{*}{\begin{sideways}ETTm2\end{sideways}} & 96    & 0.080 (0.001) & 0.212 (0.001) \\
      & 192   & 0.150 (0.024) & 0.302 (0.026) \\
      & 336   & 0.175 (0.012) & 0.334 (0.014) \\
      & 720   & 0.224 (0.008) & 0.379 (0.006) \\
\midrule
\multirow{4}[2]{*}{\begin{sideways}Exchange\end{sideways}} & 96    & 0.099 (0.003) & 0.230 (0.003) \\
      & 192   & 0.223 (0.015) & 0.353 (0.009) \\
      & 336   & 0.421 (0.002) & 0.497 (0.000) \\
      & 720   & 1.114 (0.049) & 0.807 (0.016) \\
\bottomrule
\end{tabular}%
    \end{subtable} 
    \label{tab:sd}
\end{table}

\newpage
\section{Computational Efficiency}
\label{app:efficiency}
\begin{figure}[H]
\centering
\begin{subfigure}[b]{0.49\textwidth}
\centering
\includegraphics[width=0.49\textwidth]{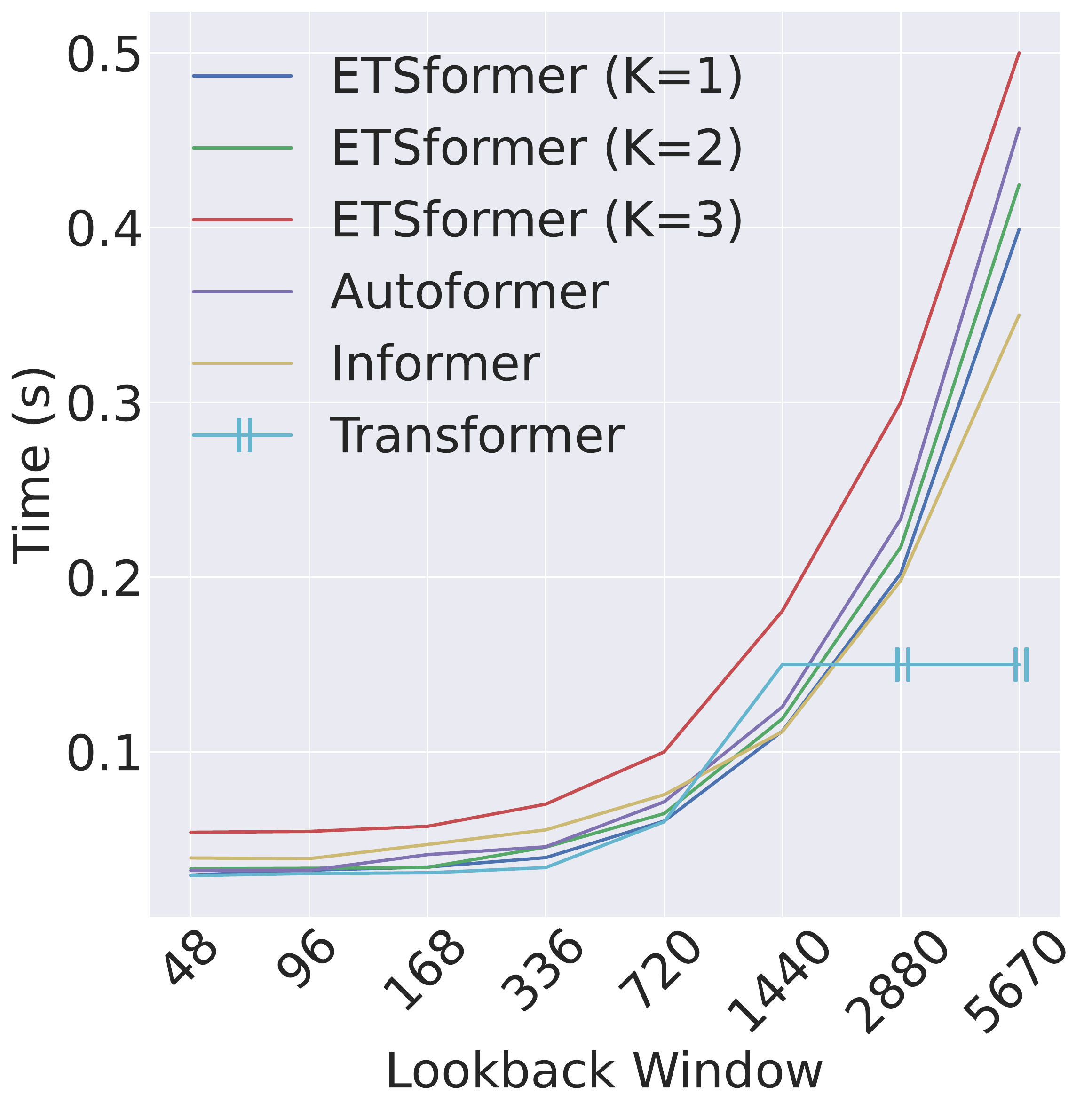}
\includegraphics[width=0.49\textwidth]{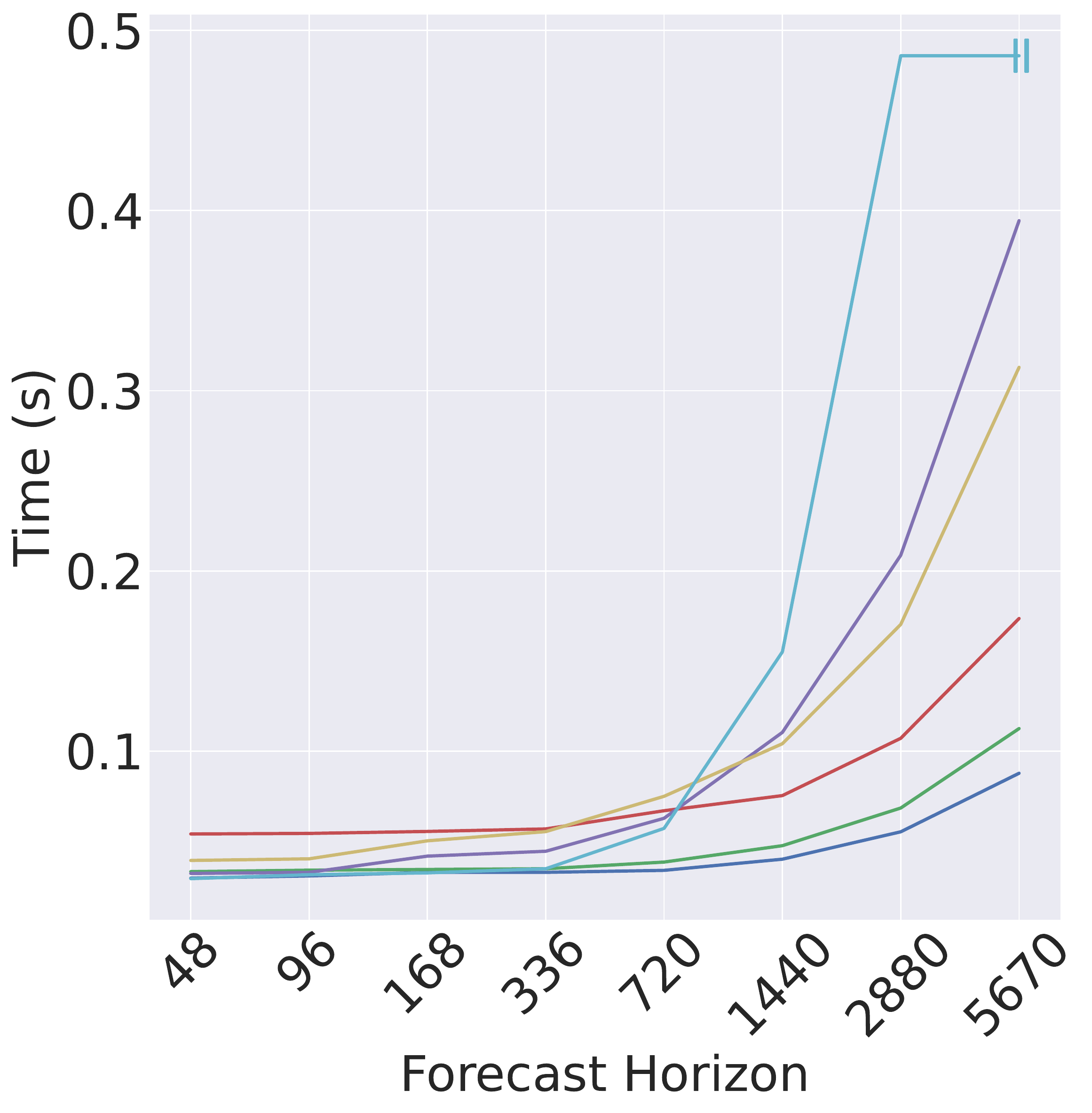}
\caption{Runtime Efficiency Analysis}
\end{subfigure}
\begin{subfigure}[b]{0.49\textwidth}
\centering
\includegraphics[width=0.49\textwidth]{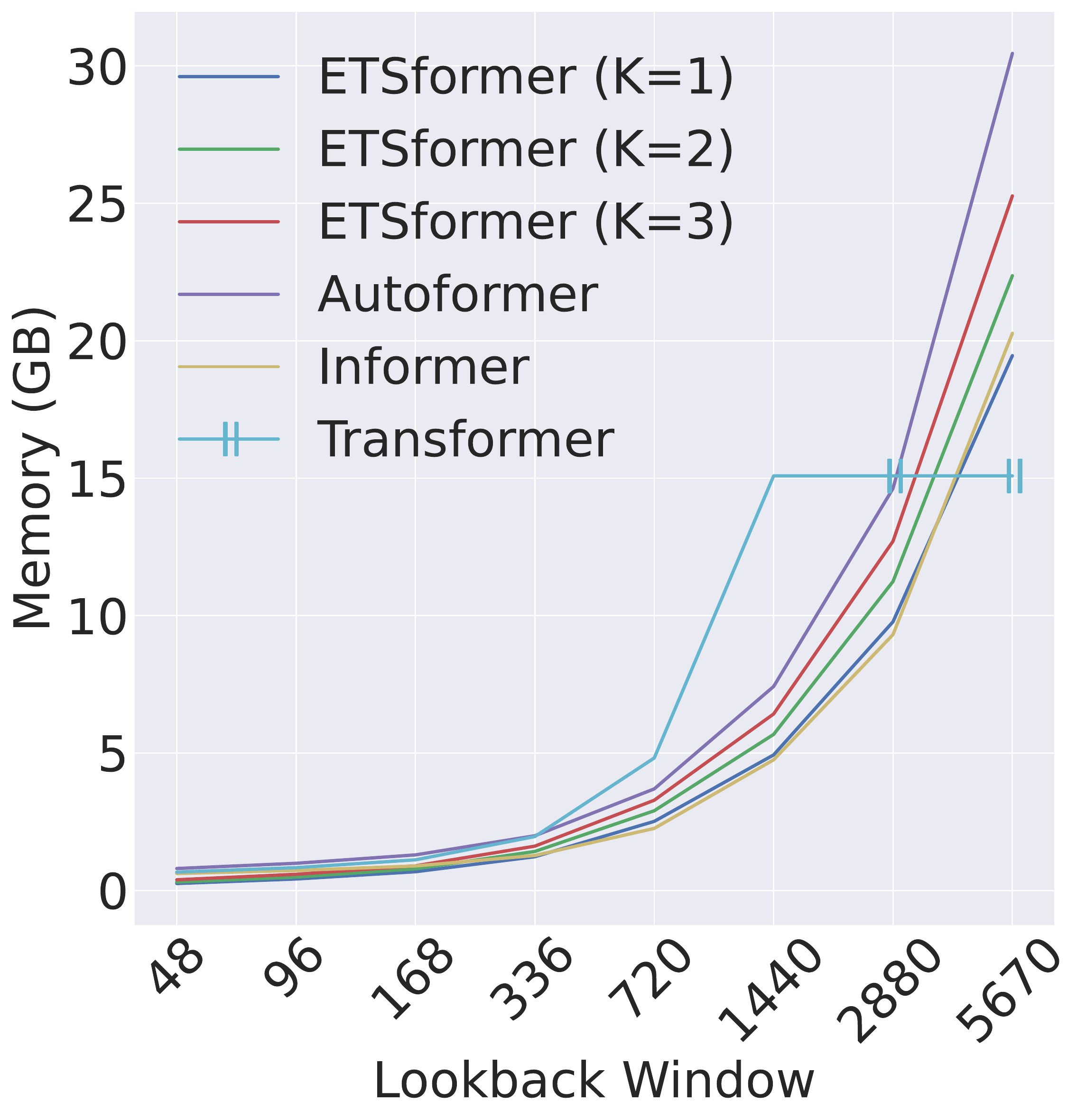}
\includegraphics[width=0.49\textwidth]{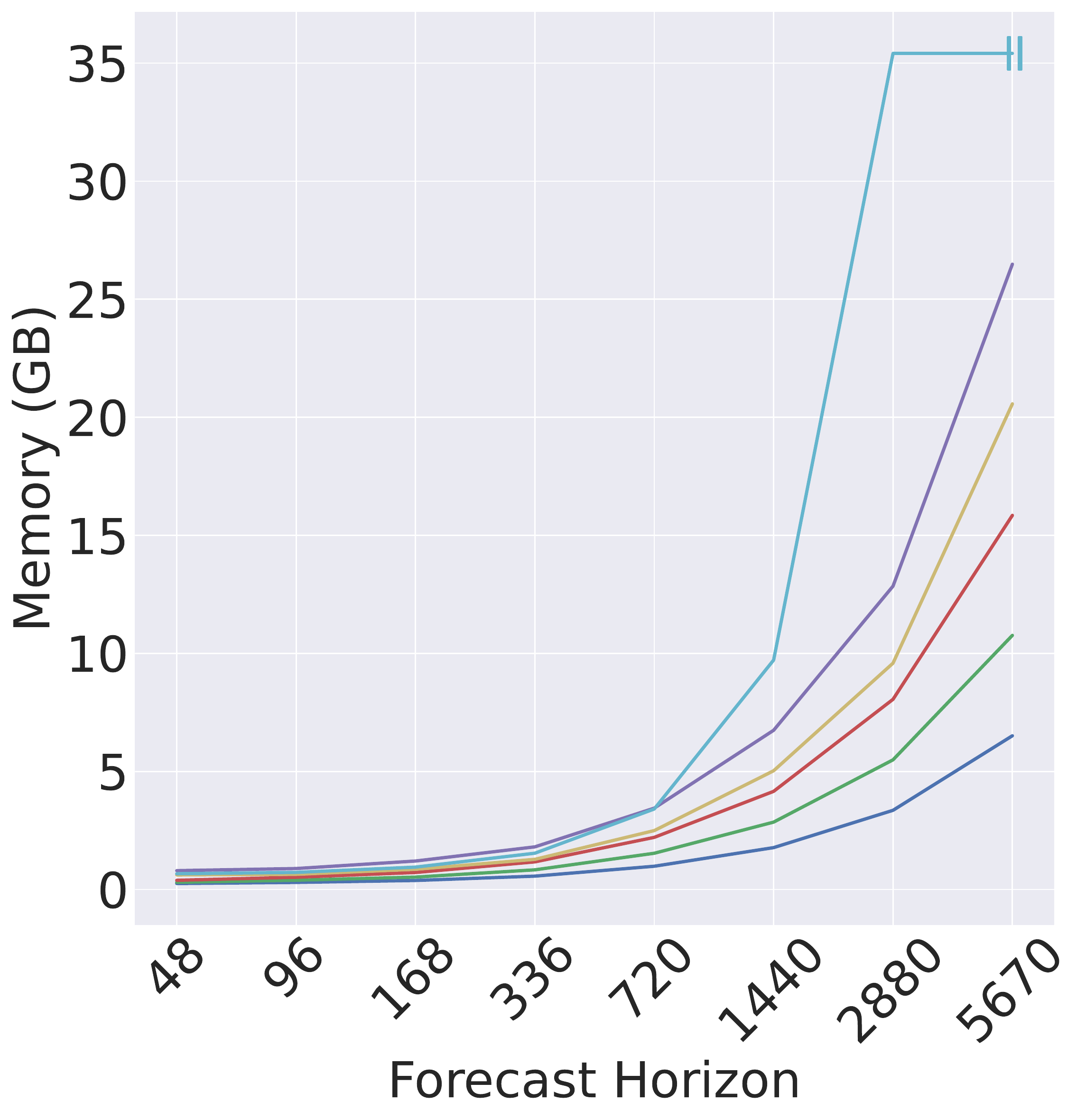}
\caption{Memory Efficiency Analysis}
\end{subfigure}
\caption{Computational Efficiency Analysis. 
Values reported are based on the training phase of ETTm2 multivariate setting. 
Horizon is fixed to 48 for lookback window plots, and lookback is fixed to 48 for forecast horizon plots.
For runtime efficiency, values refer to the time for one iteration. 
The ``\(\doublebar\)" marker indicates an out-of-memory error for those settings.}
\label{fig:efficiency}
\end{figure}
In this section, our goal is to compare {\shortname}'s computational efficiency with that of competing Transformer-based approaches. Visualized in \cref{fig:efficiency}, {\shortname} maintains competitive efficiency with compting quasilinear complexity Transformers, while obtaining state-of-the-art performance. 
Furthermore, due to {\shortname}'s unique decoder architecture which relies on its Trend Damping and Frequency Attention modules rather than output embeddings, {\shortname} maintains superior efficiency as forecast horizon increases.

\end{document}